\pdfoutput=1

\documentclass[11pt]{article}

\usepackage{EACL2023}

\usepackage{times}
\usepackage{latexsym}

\usepackage[T1]{fontenc}

\usepackage{mathtools}
\usepackage{amsmath}
\usepackage{amssymb}
\usepackage{enumerate}
\usepackage{xspace}
\usepackage[font=small]{caption} 
\usepackage{cleveref}
\usepackage{booktabs}
\usepackage{graphicx}
\usepackage{multirow}
\usepackage{hyperref}
\usepackage{subcaption}
\usepackage{caption}
\usepackage[ruled,linesnumbered]{algorithm2e}

\usepackage[utf8]{inputenc}

\usepackage{microtype}

\DeclareMathAlphabet{\mathbbb}{U}{bbold}{m}{n} %

\DeclareMathOperator*{\argmin}{arg\,min}
\newcommand*{\yoruba}{Yor\`ub\'a\xspace}
\DeclareMathOperator{\E}{\mathbb{E}}

\usepackage{inconsolata}

\title{Meta Self-Refinement for Robust Learning with Weak Supervision}

\author{Dawei Zhu$^{1}$, Xiaoyu Shen$^{1}$, Michael A. Hedderich$^{1,2}$, Dietrich Klakow$^{1}$ \\
$^1$\emph{Saarland University, Saarland Informatics Campus, Germany} \\
$^2$\emph{Cornell University, United States}\\
\texttt{\{dzhu,xshen,mhedderich,dietrich.klakow\}@lsv.uni-saarland.de}}

\begin{document}
\maketitle
\begin{abstract}
Training deep neural networks (DNNs) under weak supervision has attracted increasing research attention as it can significantly reduce the annotation cost. However, labels from weak supervision can be noisy, and the high capacity of DNNs enables them to easily overfit the label noise, resulting in poor generalization. Recent methods leverage self-training to build noise-resistant models, in which a teacher trained under weak supervision is used to provide highly confident labels for teaching the students. Nevertheless, the teacher derived from such frameworks may have fitted a substantial amount of noise and therefore produce incorrect pseudo-labels with high confidence, leading to severe error propagation. In this work, we propose Meta Self-Refinement (MSR), a noise-resistant learning framework, to effectively combat label noise from weak supervision. Instead of relying on a fixed teacher trained with noisy labels, we encourage the teacher to refine its pseudo-labels. At each training step, MSR performs a meta gradient descent on the current mini-batch to maximize the student performance on a clean validation set. Extensive experimentation on eight NLP benchmarks demonstrates that MSR is robust against label noise in all settings and outperforms state-of-the-art methods by up to 11.4\% in accuracy and 9.26\% in F1 score.
\end{abstract}

\section{Introduction}
Fine-tuning pre-trained language models has led to great success across NLP tasks. Nonetheless, it still requires a substantial amount of manual labels to achieve satisfying performance on many tasks. In reality, obtaining large amounts of high-quality labels is costly and labor-intensive \cite{DBLP:journals/biodb/DavisWRKLLSJKGHMEM13}. For certain domains, it is even infeasible due to legal issues and lack of data or domain experts. Weak supervision is a widely-used approach for reudcing such cost by leveraging labels from weak sources, e,g., heuristic rules, knowledge bases or lower-quality inexpensive crowdsourcing \cite{snorkel_2017, zhou2020_nero, lison2020_ner}. It has raised increasing attention in recent years, and efforts have been made to quantify the progress on weakly supervised learning, like the WRENCH benchmark~\cite{wrench_benchmark_2021}.

Although weak labels are inexpensive to obtain, they are often noisy and inherit biases from weak sources. Training neural networks with weak labels is challenging because of their immense capacity, which leads them to heavily overfit to the noise distribution, resulting in inferior generalization performance~\cite{DBLP:conf/iclr/ZhangBHRV17}. Various approaches have been proposed to tackle this challenge. Earlier research focused primarily on simulated noise~\cite{DBLP:conf/icassp/BekkerG16, DBLP:conf/nips/HendrycksMWG18}, required prior knowledge~\cite{denoise_2020,implyloss_2020} or relied on context-free aggregation rules without leveraging modern pre-trained language models \cite{snorkel_2017, fu2020fast_flying_squid}. 

Recently, \citet{cosine2021} proposed a contrastive regularized self-training framework that achieved state-of-the-art (SOTA) performance in several NLP tasks from the WRENCH benchmark. It trains a teacher network on weak labels, then iteratively applies the teacher to produce pseudo-labels for training a new student model. To prevent error propagation, it filters the pseudo-labels with the model confidence scores and adds contrastive feature regularization to enforce more distinguishable representations. However, we find that this approach is \textit{effective on easy tasks but fragile on challenging ones}, where the initial teacher model already have memorized a substantial amount of biases with high confidence. Consequently, confidence-based filtering is misleading and all future students will be reinforced with these initial wrong biases from the teacher. 

To address this weakness, one strategy is learning to reweight the pseudo-labels with meta learning~\cite{l2r_18,meta_weight19,wang2020adaptive}. By this means, sample weights are dynamically adjusted to minimize the validation loss instead of prefixed with potentially misleading confidence scores. Nevertheless, if the initial teacher is weak and mostly produces incorrect pseudo-labels, simply reweighting the labels does not suffice to extract enough useful training signals.

In this paper, we propose Meta Self-Refinement (MSR) to go one step further. The teacher is jointly trained with a meta objective such that the student, after one gradient step, can achieve better performance on the validation set. In each training step, a copy of the current student performs one step of gradient descent based on the teacher predictions. The teacher will then update itself towards the gradient direction that minimizes the validation loss of the student. Finally, the actual student is trained by the updated teacher. In MSR, teacher's predictions are iteratively \textit{refined}, instead of only ``reweighted”, based on the meta objective. This will enable more efficient data utilization since the teacher still has the opportunity to refine itself to provide the proper training signal, even if its initial output label is wrong. To further stabilize the training, we enhance our framework with confidence filtering when teaching the student and apply a linearly scaled learning rate scheduler to the teacher.

In summary, the main contributions are as follows: \textbf{1)} We propose a meta-learning based self-refinement framework, MSR, that allows robust learning with label noise induced by weak supervision. \textbf{2)} We analyze and quantify how label noise impacts model predictions and representation learning. We find existing methods become less effective in challenging cases when the label noise can be easily fitted. In contrast, MSR is more stable and learns better representation. \textbf{3)} Extensive experiments demonstrate that MSR consistently reduces the negative impact of the label noise, matching or outperforming SOTAs on six sequence classification and two sequence labeling tasks.\footnote{Code is available on: \url{https://github.com/uds-lsv/msr}}

\section{Related Work}
\paragraph{Learning with Noisy Labels.} Learning in the presence of label noise is a long-standing problem \cite{angluin1988learning}. \citet{DBLP:conf/iclr/ZhangBHRV17} show that deep neural networks can memorize arbitrary noise during training, resulting in poor generalization. Noise-handling techniques - by modeling~\cite{goldbergerB17, patriniRMNQ17, DBLP:conf/nips/HendrycksMWG18} or filtering~\cite{DBLP:conf/nips/HanYYNXHTS18, liSH20} the noisy instances - are proposed to conquer the label noise. While being effective, they typically assume that the noise is feature-independent which may oversimplify the noise generation process in realistic settings \cite{gu2021instance,DBLP:conf/acl-insights/ZhuHZAK22}. Recently, realistic and feature-dependent noise induced by weak supervision has received significant attention. To handle this type of noise, \citet{implyloss_2020} propose an implication loss that jointly denoises the noisy labels and weak sources. \citet{denoise_2020} denoise the weak label by considering the reliability of different weak sources and aggregating them into one cleaned label. \citet{wrench_benchmark_2021} release a benchmark, WRENCH, including various weakly supervised datasets in both text and image domains.

\paragraph{Self-Training.} Self-training \cite{yarowsky95_selftrain, self-training_lee2013pseudo} is a simple yet effective framework that is commonly used in semi-supervised learning (SSL). It typically trains a teacher model to provide pseudo-labels for the student model. Different methods have been proposed for better generalization \cite{xieLHL20, zophGLCLC020, mukherjee-awadallah-2020-ust}. Recently, self-training has been adopted to tackle weak supervision. \citet{astra_karamanolakis2021} train a teacher network that aggregates weak labels to form high-quality pseudo-labels for the student. \citet{liang2020bond, cosine2021} initialize the teacher model by training a classifier directly on the weak labels, they apply early stopping to prevent this initial teacher from memorizing the label noise. The student is then trained on the highly confident pseudo-labels provided by the teacher. While the core assumption of self-training - that highly confident pseudo-labels are reliable - is generally valid in SSL, it may not be true for feature-dependent noise induced by weak supervision, especially when the noise is easy to learn. In this case, self-training inevitably suffers more from error propagation and fails to train robust models.

\paragraph{Meta-Learning.} Recently, different works leveraged meta-learning techniques to develop noise-robust learning frameworks. The idea is to optimize an outer learner (e.g., sample weights) that guides the inner learner (the classifier) to generalize well. Often, a clean validation dataset is used as a proxy for estimating the generalization performance. \citet{l2r_18} attempt to down weight training samples that increase the validation loss. \citet{meta_weight19} employ a neural network to infer such sample weights and show a significant boost on performance under feature-independent noise. \citet{wang2020adaptive} reweight the training samples by their pseudo-labels instead of the original noisy labels. In this work, we aim to leverage meta-learning in a more flexible manner by refining the pseudo-labels instead of reweighting them. Approach-wise, the most related works are \cite{Pham21_meta_pseudo_labels, zhou2022meta} used for semi-supervised learning and model distillation, which also refine the teacher's parameters based on the student feedback. However, they work with samples from clean distributions, while we anticipate the noise memorization effect and enhance our framework with teacher warm-up and confidence filtering to suppress the error propagation.

\begin{figure}[]
    \centering
    \includegraphics[width=0.99\columnwidth]{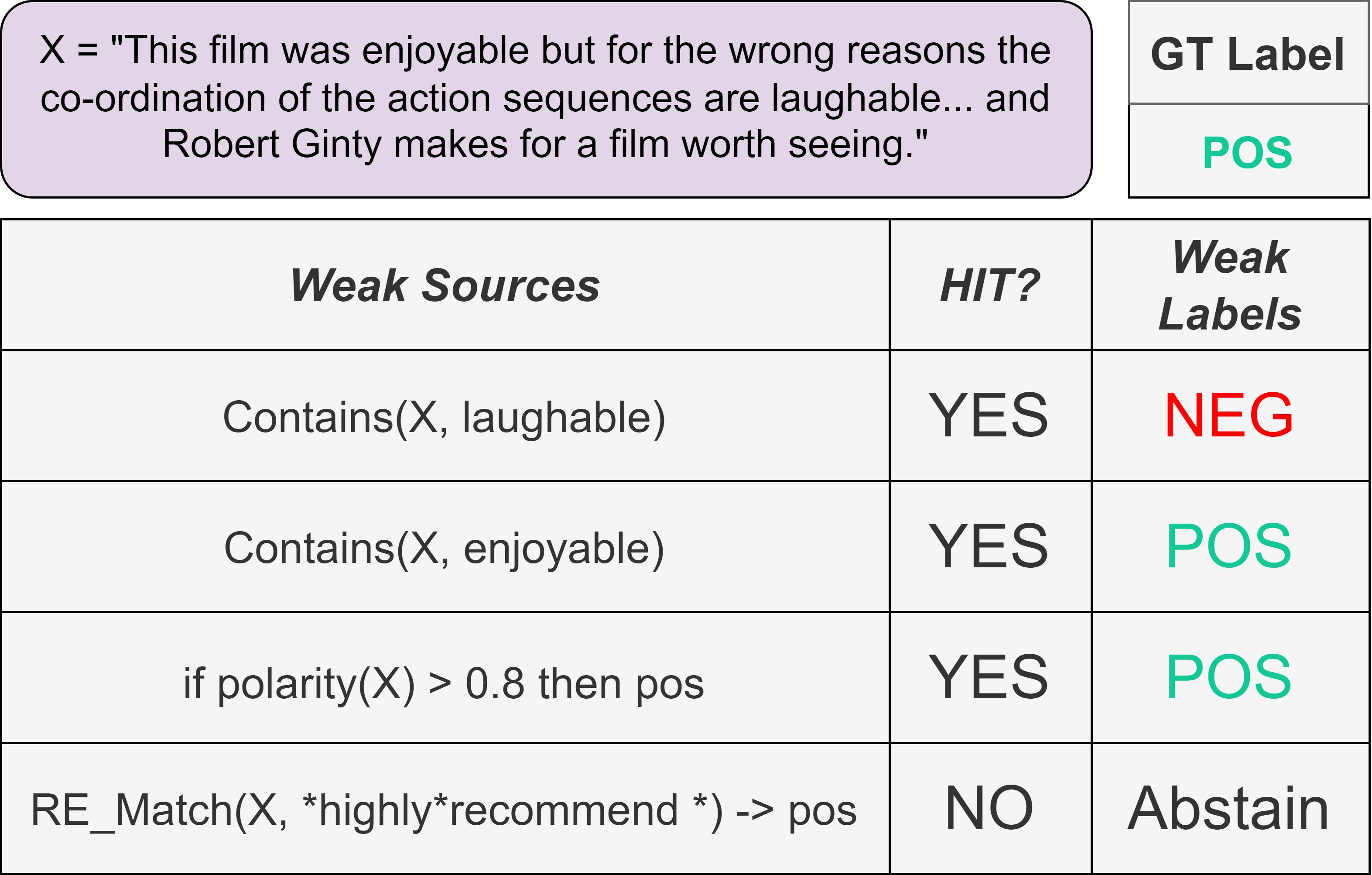}
    \caption{Sentiment analysis dataset annotated with rule-based weak sources. A weak source is triggered if a specific textual pattern is matched, after which a pre-defined label is then assigned. Otherwise, it abstains. Depending on how many weak sources are triggered, a text may obtain zero, one, or multiple weak labels.}
    \label{fig:introduction}
\end{figure}

\section{Problem Formulation}
Let $\mathcal{X}$ and $\mathcal{Y}$ be the feature and label space, respectively. 
In standard supervised learning, one is given a clean dataset $\mathcal{D}_{c}=\{(x_i,y_i)\}_{i=1}^{N}$, where $N$ is the number of samples. The clean labels $y_i$ are supposed to be annotated by human experts.

In weak supervision, a dataset is labeled by weak sources rather than humans. Weak sources can have diverse forms like lexical rules, knowledge bases, pre-trained models, lower-quality inexpensive crowdsourcing, etc. Figure \ref{fig:introduction} shows an example of text labeled via weak supervision. Compared to manual annotations, weak labels contain more mistakes.
We denote the dataset labeled by weak sources by $\mathcal{D}_{w}=\{(x_i,\hat{y}_i)\}_{i=1}^{N}$ where $\hat{y}_{i}$ is the weak label.\footnote{Multiple weak sources may be triggered simultaneously by a sample. In this case, we can use different aggregation methods like majority voting to determine the final weak label.} Since weak sources might not cover all data, we may have a set of unlabeled data $\mathcal{D}_u$ in addition to $\mathcal{D}_w$. We use $\mathcal{D}_a = \mathcal{D}_w \cup \mathcal{D}_u$ to denote the full set of data. Moreover, as we do not make any assumption on the quality of the weak labels, their distribution can deviate arbitrarily from the distribution of clean labels. Learning with only weak labels can lead to unbounded model errors~\cite{DBLP:journals/corr/MenonRN16, gu2021instance}. Hence, following standard practice in weak supervision, we assume the access to a small clean validation set $\mathcal{D}_{v}=\{(x^{v}_i,y^{v}_i)\}_{i=1}^{M}$ where $M \ll N$. $\mathcal{D}_{v}$ is used for early stopping, hyper-parameter tuning or meta-learning so that the learned model will not fully overfit the noisy weak labels~\cite{l2r_18,meta_weight19,wrench_benchmark_2021}.

\section{Meta Self-Refinement}
\begin{figure*}[]
    \centering
    \includegraphics[width=1.0\textwidth]{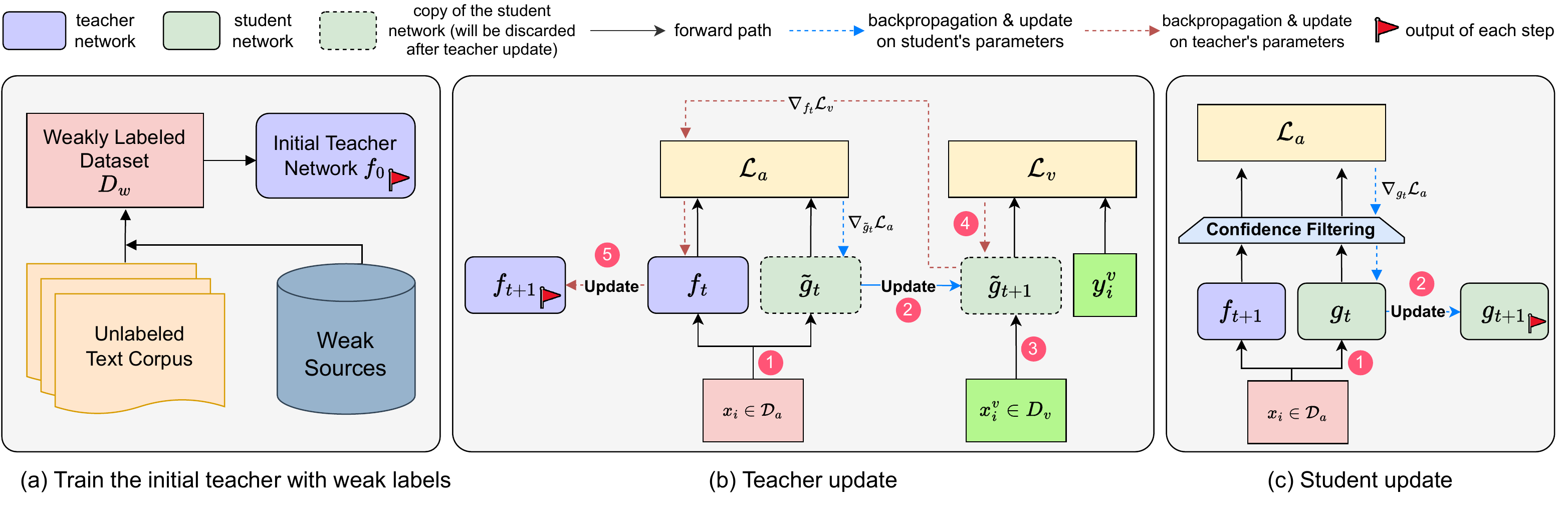}
    \caption{Illustration of our proposed Meta-Self Refinement method (MSR). (a) We start by fine-tuning a PLM on weak labels with early stopping, which yields an initial teacher $f_1$. (b) At each training step $t$, $f_t$ gets training signals by performing a ``teaching experiment'' on $\tilde{g}_t$: a copy of the student network $g_t$. $\tilde{g}_t$ is updated by fitting $f_t$ with the loss function $\mathcal{L}_a$. $f_t$ is then updated to minimize the validation loss $\mathcal{L}_v$ of $\tilde{g}_{t+1}$. (c): $g_t$ is updated by fitting $f_{t+1}$ with confidence filtering under the loss $\mathcal{L}_a$.}
    \label{fig:method_visualization}
\end{figure*}

We propose a novel meta-learning based framework, named Meta Self-Refinement (MSR), to tackle the label noise induced by weak supervision. In contrast to conventional self-training methods, where the teacher model is fixed after being trained on weakly labeled data, MSR enables the teacher to refine itself based on student performance on the clean validation set, yielding higher-quality labels and more accurate confidence estimates. In this section, we first provide an overview of its training objective (\cref{sec:objective}), then go into the training details (\cref{sec:details}). Figure~\ref{fig:method_visualization} illustrates the full training process.

\subsection{Training Objective}
\label{sec:objective}
MSR contains a teacher network $f$ and a student network $g$, both are functions that map $\mathcal{X} \rightarrow \mathcal{Y}$. $f$ is initialized by fine-tuning a pre-trained language model (PLM) on the weakly labeled data $\mathcal{D}_{w}$:
\begin{equation}
f_1 = \argmin_{f}  \E_{(x_{i}, \hat{y}_{i}) \in \mathcal{D}_{w}} \mathcal{L}(\hat{y}_{i}, f(x_{i}))
\label{eq:teacher_init}
\end{equation}
where $\mathcal{L}$ denotes the loss function. We use the cross entropy loss throughout the paper:
\begin{equation}
\mathcal{L}(p,q) = -\E_{y \sim p(y)} \log q(y)
\label{eq:loss}
\end{equation}
$p$ and $q$ are distributions over the label space $\mathcal{Y}$. The initial student network, $g_1$, is the PLM without fine-tuning on any data.

In conventional self-training, $f_1$ is used to provide pseudo-labels to train the student. By selecting higher-quality pseudo-labels via confidence filtering~\cite{cosine2021} or uncertainty estimation~\cite{mukherjee-awadallah-2020-ust}, the student can often outperform its teacher. However, as the teacher is trained solely on the weak labels, it can easily inherit unexpected biases and provide misleading signals to the student. In MSR, instead of using a fixed teacher to provide pseudo-labels, we use student performance on the clean validation set as a feedback signal to dynamically refine the teacher. Specifically, the objective for the teacher $f$, formulated as in Equation \ref{eq:objective_teacher}, is that \emph{the student network, after fitting the teacher's output labels on $\mathcal{D}_a$, can perform best on the validation set $\mathcal{D}_{v}$}:
\begin{equation}
\label{eq:objective_teacher}
\begin{split}
&f^\star = \argmin_{f}\E_{(x^{v}_{i}, y^{v}_{i}) \in \mathcal{D}_{v}} \mathcal{L}(y^{v}_{i}, g_{f}^\prime(x^{v}_i))\\
 &g_{f}^\prime = \argmin_{g}\E_{x_{i} \in \mathcal{D}_{a}} \mathcal{L}(f(x_{i}), g(x_{i}))
\end{split}
\end{equation}
where $g^\prime$ is the student network after fitting output labels from $f$ on $\mathcal{D}_a$. 
Intuitively, MSR aims to find the best teacher to help the student achieve the lowest validation loss. After finding the optimal teacher $f^\star$ in Equation \ref{eq:objective_teacher}, the student can then be obtained by learning from the output labels of $f^\star$:
\begin{equation}
g^\star = \argmin_{g}\E_{x_{i} \in \mathcal{D}_{a}} \mathcal{L}(f^\star(x_{i}), g(x_{i}))
\label{eq:objective_student}
\end{equation}

\begin{algorithm}
\small
\caption{MSR Algorithm}\label{algorithm}
\KwIn{Initial teacher network $f_{1}$ trained according to Eq. \ref{eq:teacher_init}. Student network $g_{1}$, number of training steps $T$, teacher's learning rate scheduler $R(t)$, confidence threshold $\tau$, $\mathcal{D}_{a}$, $\mathcal{D}_{v}$.}
\KwResult{$f_{T}$, $g_{T}$}
\For{$t \gets 1 \dots T$}{
$\{x_{i}\} \gets$ \text{SampleMiniBatch}( $\mathcal{D}_{a}$)\\
$\{x_{i}^{v}, y_{i}^{v}\} \gets$ \text{SampleMiniBatch}( $\mathcal{D}_{v}$)\\
\tcp{Teacher Update}
$\tilde{g}_{t} \gets \text{Copy}(g_t) $\\
$\tilde{g}_{t+1} \gets \tilde{g}_{t} - \lambda_{s} \E_{x_i}\nabla_{\tilde{g}_{t}} \mathcal{L}(f_t(x_i), \tilde{g}_t(x_i))$\\
$f_{t+1} \gets f_{t} - R(t) \E_{(x_{i}^{v}, y_i^{v})}\nabla_{f_{t}}\mathcal{L}(y_i^{v},\tilde{g}_{t+1}(x_{i}^{v}))$\\
\tcp{Student Update}
$w(f_{t+1}(x_i)) \gets \mathbbb{1}( 1 - \frac{H(f_{t+1}(x_i))}{\log(k)} \geq  \tau)$\\
$g_{t+1} \gets g_{t} - \lambda_{s} \E_{x_i}\nabla_{g_{t}}w(f_{t+1}(x_i))\mathcal{L}(f_{t+1}(x_i),g_t(x_i))$\\
}
\end{algorithm}

\subsection{Training Details}
\label{sec:details}

Finding the exact $f^\star$ in Equation \ref{eq:objective_teacher} involves solving two nested loops of optimization, and each loop can be computationally expensive given the large size of $\mathcal{D}_a$. We resort to an online approximation to merge Equation \ref{eq:objective_teacher} and \ref{eq:objective_student} into an iterative training pipeline. At each training step $t$, the teacher $f_t$ is first updated based on the meta-objective of ``learning to teach'', the student $g_t$ is then trained by the updated teacher. 

\paragraph{Teacher Update.} To update the teacher in an efficient way, we approximate the inner loop in Equation \ref{eq:objective_teacher} with a single-step gradient descent of the student network. Namely, the objective of the teacher is changed so that the current student, after \emph{one single gradient descent step} of fitting the teacher, can perform best on the validation set. To do so, the teacher will first conduct a ``teaching experiment'' on a copy of the current student, denoted as $\tilde{g}_{t}$. $\tilde{g}_{t}$ is updated for one gradient descent step to fit the teacher's pseudo labels\footnote{We use SGD for illustration purposes. The AdamW \cite{LoshchilovH19_adamW} optimizer is used in our experiments.}:
\begin{equation*}
    \tilde{g}_{t+1} = \tilde{g}_{t} - \lambda_{s} \E_{x_i \sim \mathcal{D}_{a}}\nabla_{\tilde{g}_{t}} \mathcal{L}(f_t(x_i), \tilde{g}_t(x_i))
    \label{eq:student_update}
\end{equation*}
where $\lambda_{s}$ is the learning rate of the student network. Afterwards, we update the teacher network to minimize the validation loss of $\tilde{g}_{t+1}$:
\begin{equation*}
    f_{t+1} = f_{t} - \lambda_{t} \E_{(x_i^v, y_i^v) \sim \mathcal{D}_{v}}\nabla_{f_{t}}\mathcal{L}(y_i^v,\tilde{g}_{t+1}(x_i^v))
    \label{eq:teacher_update}
\end{equation*}
where $\lambda_{t}$ is the learning rate of the teacher network.
It requires calculating second derivatives over $f_t$. We always use soft labels from the teacher for $\mathcal{L}(f_t(x_i), \tilde{g}_t(x_i))$, so the whole process is fully differentiable. Note that $\tilde{g}_{t}$ is only used in the ``teaching experiment'' to help update the teacher. It will be discarded after the teacher is updated.

\paragraph{Student Update.} After obtaining $f_{t+1}$, the real student network is updated with the same objective as in Equation \ref{eq:objective_student}, except that we use the updated teacher $f_{t+1}$ instead of $f^\star$. As the teacher has performed the ``teaching experiment'', it will provide more useful signals to guide the student.\footnote{In theory, if the teacher network is strong enough to generalize among different batches, we can directly update the real student in the ``teaching experiment'', in the hope that the teacher from the last step can also work in the current batch. However, in practice, we find this mismatch leads to poor performance.}

\paragraph{Teacher Learning Rate Scheduler.}
We find the teacher is rather sensitive to its learning rate in practice. If the learning rate is large from the start, the teacher may over-adjust itself due to the large performance gap between the teacher and the student. If the learning rate is small, the teacher will adjust itself too slowly so that more noisy pseudo-labels are passed to the student network. Therefore, we apply a linear learning rate scheduler $R(t)= \frac{t\lambda_{t}}{T}$ to the teacher network where $t$ denotes the current iteration and $\lambda_{t}$ is the targeted learning rate for the teacher. By this means, the teacher's learning rate will gradually increase as it gets better at teaching.

\paragraph{Confidence-Based Label Filtering.}
Despite having the opportunity to refine itself, the teacher inevitably produces some wrong pseudo labels during training, especially at early iterations of self-refinement. To further reduce error propagation, we only select labels with high confidence to guide the student model. The student is updated as follows:
\begin{equation*}
\label{eq:filter_student}
\begin{split}
g_{t+1} = g_{t} &- \lambda_{s} \E_{x_i \sim \mathcal{D}_{a}}\nabla_{g_{t}} \mathcal{L}(f_{t+1}(x_i),g_t(x_i))\\
 &\times \mathbbb{1}( 1 - \frac{H(f_{t+1}(x_i))}{\log(k)} \geq  \tau)
\end{split}
\end{equation*}
where $\mathbbb{1}$ is the indicator function, $H(f_{t+1}(x_i))$ is the entropy of the distribution $f_{t+1}(x_i)$, $k$ is the number of classes in $\mathcal{Y}$ and $\tau$ is a pre-defined confidence threshold. $\log(k)$ is the upper bound of the entropy for $k$-classification tasks. By this means, only low-entropy (high-confidence) predictions from the teacher are learned. Note that the filtering strategy is only applied to the actual student update step, not during the teaching experiment. Otherwise, the teacher will ignore low-confident samples as they do not contribute to teacher update.

Putting all together, Algorithm~\ref{algorithm} summarizes the self-refinement process.

\section{Experimental Settings}
\paragraph{Datasets.} WRENCH \cite{wrench_benchmark_2021} is a well-established benchmark for weak supervision and offers weak labels for various datasets. We compare different baselines on six NLP datasets from WRENCH including both sequence classification and Named-Entity Recognition (NER) tasks. For sequence classification, we include AGNews \cite{agnews_zhang2015character}, IMDB \cite{imdb_maas-etal-2011-learning}, Yelp \cite{agnews_zhang2015character}, and TREC \cite{trec_li2002learning}. For NER tasks, CoNLL-03 \cite{conll03_sang2003introduction} and OntoNotes 5.0 \cite{ontonotes_pradhan-etal-2013-towards} are used.
In addition, we further include two sequence classification datasets in low-resource languages, \yoruba and Hausa \cite{DBLP:conf/emnlp/HedderichAZAMK20}, to involve evaluation cases in diverse languages. Table~\ref{tab:dataset} summarizes the basic statistics of the datasets. Majority voting over weak sources is used to determine a single label for each sample.

\begin{table}[th]
\centering
	\resizebox{\columnwidth}{!}{
		\begin{tabular}{cccccccc}
			\toprule \bf Dataset & \bf Task & \bf \# Class  &\bf \# Train & \bf \# Val & \bf \# Test \\ \midrule
            AGNews &Topic &4 &96,000 & 12,000 & 12,000  \\ %
            IMDB &Sentiment &2 &20,000 & 2,500 & 2,500 \\ %
            Yelp &Sentiment& 2 &30,400 & 3,800 & 3,800  \\ %
            TREC & Question & 6 & 4,965 & 500 & 500   \\ %
            \yoruba &Topic &7 &1,340 & 189 & 379  \\ %
            Hausa &Topic &5 &2,045 & 290 & 582  \\ %
            CoNLL03 & NER & 4 & 14,041 & 3,250 & 3,453     \\
            OntoNotes5.0 & NER & 18 &  115,812 & 5,000 &  22,897  \\
            \bottomrule
		\end{tabular}
	}%
	\caption{Dataset statistics. Refer to Appendix \ref{sec:appendix_dataset_details} for more details on datasets.}
	\label{tab:dataset}

\end{table}

\paragraph{Implementation.}
RoBERTa-base~\cite{liu19roberta} is used as the PLM for English datasets and multilingual BERT-base~\cite{devlinCLT19} for non-English ones. We utilize the higher\footnote{https://github.com/facebookresearch/higher} library to perform second-order optimization. Refer to Appendix~\ref{sec:appendix_implementation_details} for detailed hyper-parameter configurations.

\begin{table*}[htb!]
	\centering
		\resizebox{\textwidth}{!}{
			\begin{tabular}{lcccccccc}
				\toprule
    \textbf{Method} &  \begin{tabular}[c]{@{}c@{}} \textbf{AGNews}\\ \textbf{(Acc)} \end{tabular}&  \begin{tabular}[c]{@{}c@{}} \textbf{IMDB}\\ \textbf{(Acc)} \end{tabular}&  \begin{tabular}[c]{@{}c@{}} \textbf{Yelp}\\ \textbf{(Acc)} \end{tabular}&  \begin{tabular}[c]{@{}c@{}} \textbf{TREC}\\ \textbf{(Acc)} \end{tabular}&  \begin{tabular}[c]{@{}c@{}} \textbf{\yoruba}\\ \textbf{(Acc)} \end{tabular}&  \begin{tabular}[c]{@{}c@{}} \textbf{Hausa}\\ \textbf{(Acc)} \end{tabular}&  \begin{tabular}[c]{@{}c@{}} \textbf{CoNLL-03}\\ \textbf{(F1)} \end{tabular}&  \begin{tabular}[c]{@{}c@{}} \textbf{OntoNotes}\\ \textbf{(F1)} \end{tabular} \\
\midrule
				\multicolumn{6}{l}{\textbf{Fully-Supervised Result}}\\
				FT-CL& $92.61$ & $93.20$ & $96.91$ & $96.67$ & $77.24$ & $81.57$ & $92.27$ & $85.74$ \\
				\hline 
				\multicolumn{6}{l}{\textbf{Label Models}}\\
				Majority  & $63.84$ & $71.04$ & $70.21$ & $60.80$  & $58.05$ & $47.93$ & $60.38$ & $58.92$\\
				Snorkel~\cite{snorkel_2017} & $62.67$ & $71.60$ & $68.92$ & $59.60$ & $62.80$ & $47.94$ & $62.88$ & $58.46$\\
				\hline 
				\multicolumn{6}{l}{\textbf{DNN Baselines}}\\
				FT-WL & $85.73_{\pm 0.43}$ & $83.43_{\pm 0.91}$ & $87.71_{\pm 1.46}$ & $66.80_{\pm 1.44}$ & $64.12_{\pm 0.83}$ & $46.13_{\pm 0.43}$ & $69.20_{\pm 0.33}$ & $67.26_{\pm 0.62}$ \\
				FT-WLST$^\dagger$~\cite{self-training_lee2013pseudo} & $88.61_{\pm 0.71}$ & $89.50_{\pm 0.65}$ & $95.32_{\pm 0.70}$ & $76.00_{\pm 2.21}$ & $67.28_{\pm 1.12}$ & $49.22_{\pm 1.39}$  &  $69.87_{\pm 0.36}$ &  $64.13_{\pm 1.45}$\\ 

				L2R~\cite{l2r_18}$^\diamond$ & $87.28_{\pm 1.00}$ & $82.76_{\pm 1.59}$ & $93.34_{\pm 0.91}$ & $83.40_{\pm 2.01}$ & $70.45_{\pm 0.69}$ & $55.67_{\pm 0.88}$ & $79.15_{\pm 1.34}$ & $70.66_{\pm 0.74}$ \\
				Meta-Weight-Net$^\diamond$~\cite{meta_weight19} & $85.96_{\pm 0.80}$ & $86.72_{\pm 0.50}$ & $86.97_{\pm 0.74}$ & $69.39_{\pm 1.27}$ & $70.00_{\pm 2.12}$ & $48.63_{\pm 0.96}$ & $69.54_{\pm 1.43}$ & $69.11_{\pm 1.20}$ \\
				Denoise~\cite{denoise_2020} & $83.45_{\pm 0.68}$ & $76.22_{\pm 0.92}$ & $71.56_{\pm 0.56}$ & $61.80_{\pm 1.30}$ & $66.10_{\pm 1.52}$ & $49.31_{\pm 0.93}$ & $72.96_{\pm 0.51}$  & $67.64_{\pm 1.06}$ \\ 
				UST$^\dagger$~\cite{mukherjee-awadallah-2020-ust} & $87.78_{\pm 0.59}$ & $86.74_{\pm 1.18}$ & $91.23_{\pm 0.90}$ & $77.20_{\pm 2.29}$ & $68.12_{\pm 0.71}$ & $47.67_{\pm 0.91}$ & $69.48_{\pm 1.69}$ & $66.98_{\pm 0.99}$ \\

				COSINE$^\dagger$~\cite{cosine2021} & $89.34_{\pm 0.76}$ & $\textbf{90.52}_{\pm 1.06}$ & $\textbf{95.48}_{\pm 0.13}$ & $82.60_{\pm 1.09}$ &$68.87_{\pm 0.82}$ & $49.66_{\pm 1.32}$ & $70.60_{\pm 0.87}$ & $64.59_{\pm 1.08}$\\
				
				\hline 
				\hline
				\multicolumn{6}{l}{\textbf{Our}  \ \textbf{Framework}}\\
				Teacher-Init ($f_{1}$) & $86.37_{\pm 0.00}$ & $85.00_{\pm 0.00}$ & $89.92_{\pm 0.00}$ & $69.00_{\pm 0.00}$ & $65.44_{\pm 0.00}$ & $46.74_{\pm 0.00}$ & $69.73_{\pm 0.00}$ & $68.25_{\pm 0.00}$ \\
				MSR$^\dagger$ $^\diamond$ & $\textbf{89.92}_{\pm 0.64}$ & $89.16_{\pm 0.91}$ & $95.00_{\pm 0.35}$ & $\textbf{94.80}_{\pm 0.29}$ & $\textbf{72.56}_{\pm 0.78}$ & $\textbf{59.11}_{\pm 0.78}$ & $\textbf{88.41}_{\pm 0.63}$ & $\textbf{74.59}_{\pm 0.84}$ \\
				\bottomrule
			\end{tabular}
		}%
		
	\caption{Accuracy and F1 score (in \%) on eight NLP tasks. The mean and standard deviation over five trials are reported. Teacher-Init is the best model checkpoint selected from the five trials of FT-WL (according to the validation performance). For a fair comparison, all self-training-based models use the same Teacher-Init checkpoint. MSR matches or outperforms SOTAs on all tasks. $^\dagger$ self-training based method. $^\diamond$ meta-learning based method. }
	\label{tab:main_result}
\end{table*}

\paragraph{Baselines.}
We compare our method with prior work on learning with noisy labels. 1) \textbf{Majority} applies majority vote on the weak labels. Ties are broken by randomly selecting a weak label. 2) \textbf{Snorkel} \cite{snorkel_2017} trains a labeling model that aggregates weak labels from different weak sources. 3) \textbf{FT-WL} fine-tunes PLMs on the weak labels. 4) \textbf{FT-WLST} further applies classic self-training \cite{self-training_lee2013pseudo} on the model obtained by FT-WL. 5) \textbf{L2R} \cite{l2r_18} uses a meta-learning framework to reweight weakly labeled samples. 6) \textbf{Meta-Weight-Net} \cite{meta_weight19} also applies meta-learning based sample reweighting. However, the weights are computed through an external reweighting network. 7) \textbf{Denoise} \cite{denoise_2020} iteratively corrects wrong annotations in the training set, and the classifier learns with the corrected labels. 8) \textbf{UST} \cite{mukherjee-awadallah-2020-ust} is a self-training based method that assigns higher weights to samples that the teacher is certain about. The uncertainties are measured via MC-dropout on the predictions \cite{mc_drop_uncertain_16}. 9) \textbf{COSINE} \cite{cosine2021} trains its student network with pseudo-labels which the teacher is highly confident about. In addition, contrastive regularization is introduced to further alleviate error propagation.

For our proposed framework, we report the performances of both \textbf{Teacher-Init ($f_1$)}: the initial teacher trained directly on weak labels, and \textbf{MSR}: the final student model ($g_{T}$). $f_1$ is obtained by running FT-WL five times and selecting the best one among them according to the validation performance. \emph{For a fair comparison, the same $f_1$ is used as the initial teacher for all self-training based models}.
Finally, we also include the results of fine-tuning PLMs on the clean versions of each dataset, denoted by \textbf{FT-CL}, to represent the upper bound performance.

\begin{figure*}[t]
    \centering
     \begin{subfigure}[b]{0.67\columnwidth}
         \centering
         \includegraphics[width=\columnwidth]{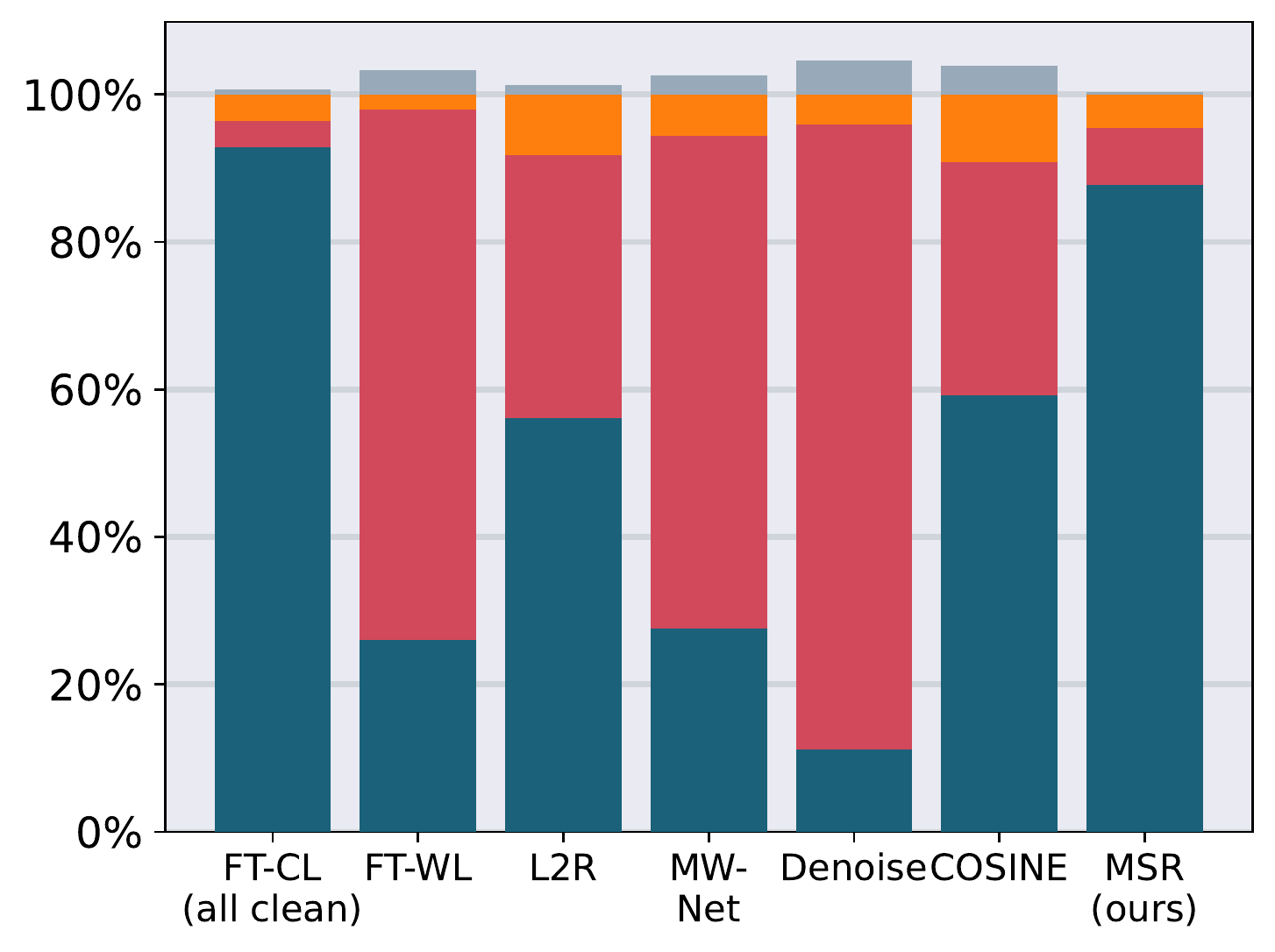}
         \caption{TREC}
     \end{subfigure}\hfill
      \begin{subfigure}[b]{0.67\columnwidth}
         \centering
         \includegraphics[width=\columnwidth]{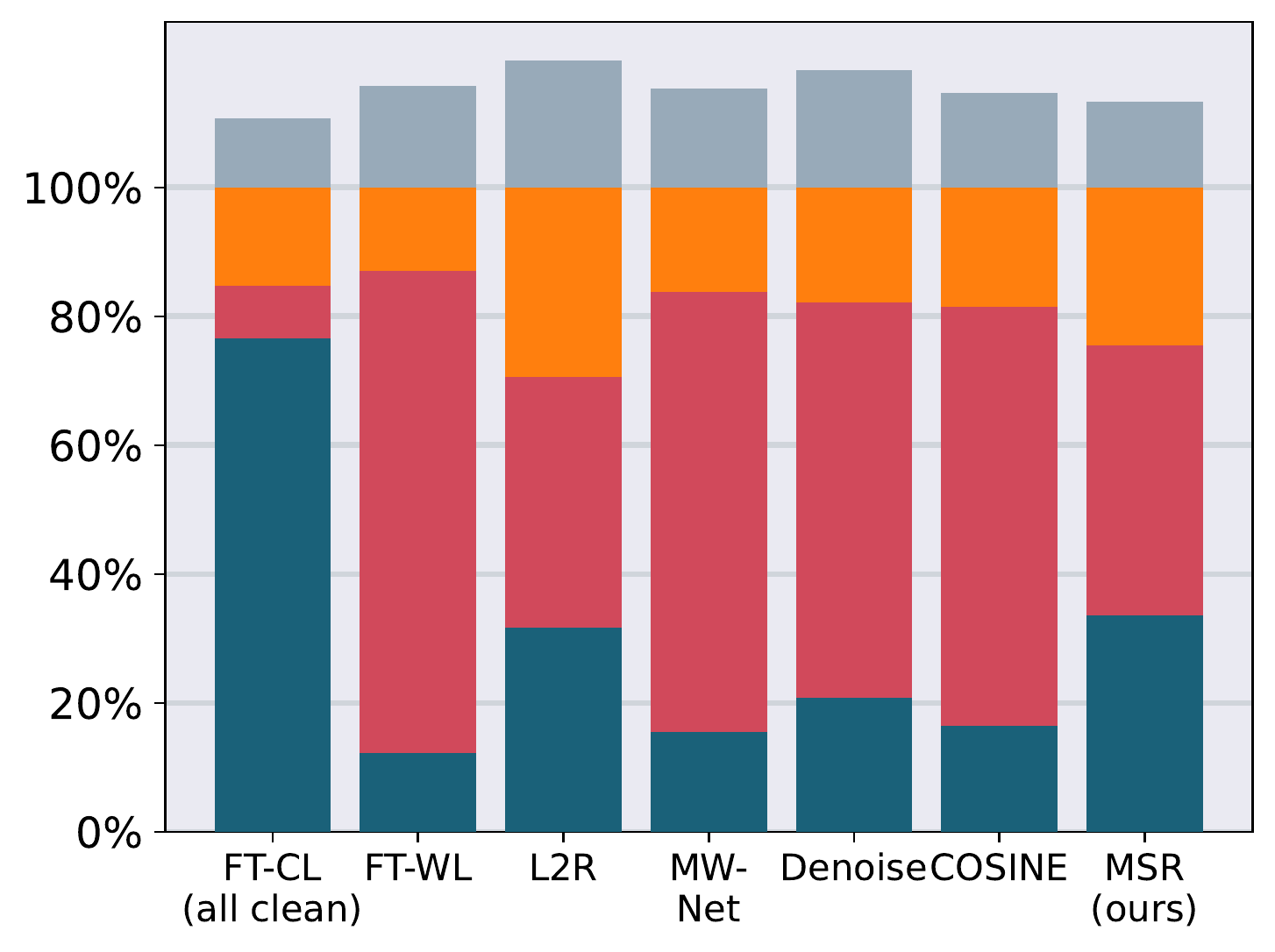}
         \caption{Hausa}
     \end{subfigure}\hfill
      \begin{subfigure}[b]{0.67\columnwidth}
         \centering
         \includegraphics[width=\columnwidth]{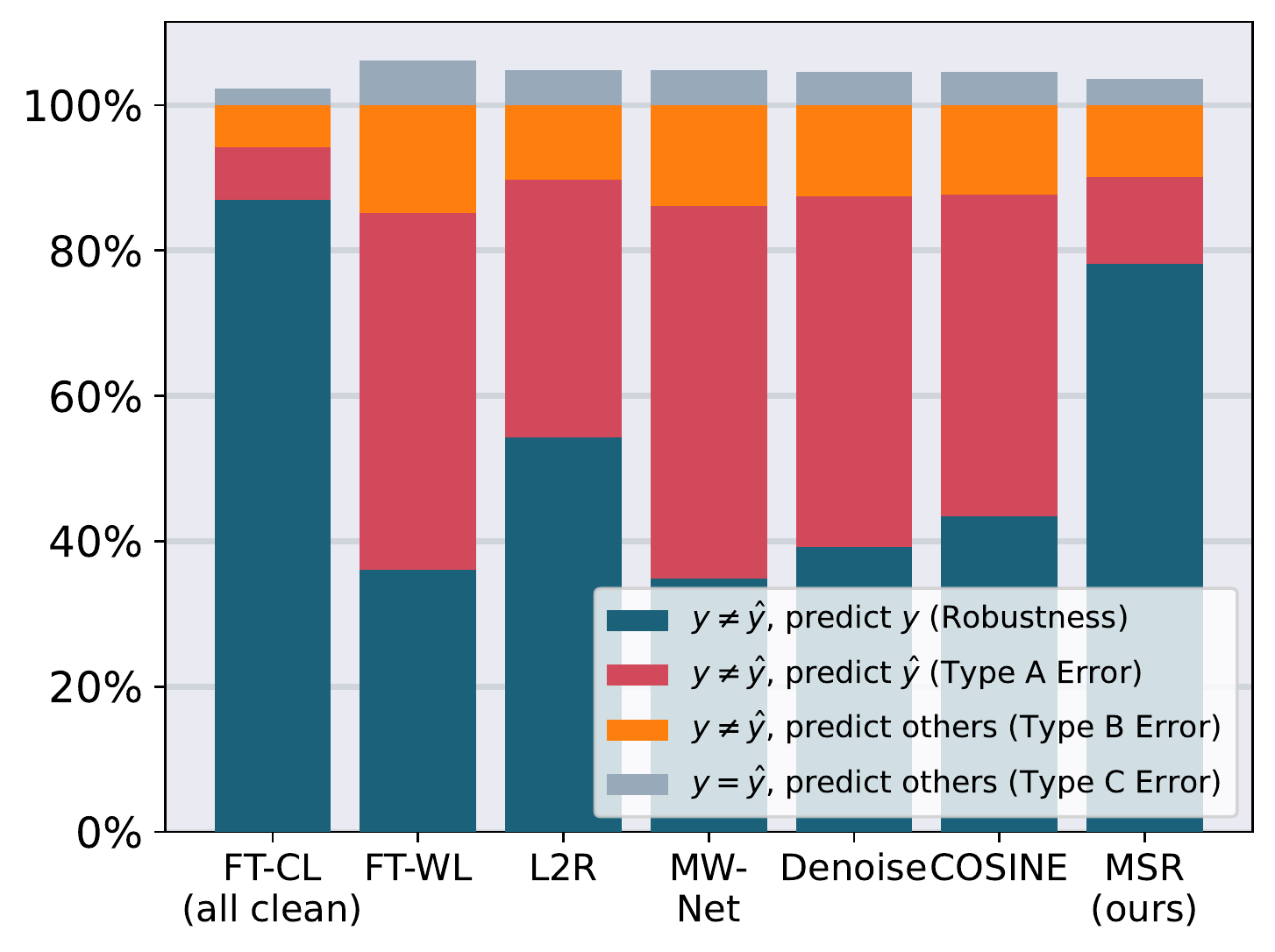}
         \caption{CoNLL-03}
     \end{subfigure}
    \caption{\small Prediction error decomposition of various weak supervision baselines, evaluated on the test sets. A model is considered robust against label noise if it manages to predict the correct labels despite the wrong weak labels (the robustness is represented by the blue bars). Otherwise, it conforms to the weak label (Type-A error) or predict another incorrect label (Type-B error), which has a negative effect on generalization. The Type-C error rate quantifies the proportion of incorrect model predictions  when weak labels are correct. MSR consistently reduces the Type-A error rate and attains a high level of noise robustness.}
    \label{fig:robustness_plot}
\end{figure*}

\section{Results}

\paragraph{Comparison with Baselines.}
\label{sec:main_result}
Table \ref{tab:main_result} shows a comparison among different methods. MSR matches or outperforms SOTAs on all eight datasets. FT-WL outperforms majority voting over the weak labels in all cases except Hausa, which leads to a minor drop. This confirms that PLMs encode useful knowledge in their parameters, enabling them to generalize better than weak rules they are trained on. This phenomenon is particularly noticeable on AGNews, IMDB, and Yelp: direct fine-tuning on the noisy labels (FT-WL) can already achieve decent performance (accuracy above 83\%). \textit{We consider them easy tasks since label noise has only a minor impact on performance of PLMs and decent generalization can be attained even without specific noise-handling.} Applying self-training to such simple tasks lead to further performance improvement. COSINE, a SOTA self-training based model, can even perform comparably to the fully supervised model on these three datasets. On the other five datasets, however, FT-WL performs poorly and conventional self-training methods provide little performance boost (even a disservice on OntoNotes). This implies that \emph{self-training relies on a well-performed initial teacher to work effectively}. On challenging datasets where the initial teacher is weak, it struggles to achieve further performance gain. Meta-learning based methods such as L2R performs better than COSINE on these challenging datasets. \emph{MSR can further boost the performance on all the challenging datasets by up to 11.4\% in accuracy or 9.26\% in F1 score while maintaining comparable results on simpler datasets}.

\begin{figure}[t]
    \centering
     \begin{subfigure}[t]{0.5\columnwidth}
         \centering
         \includegraphics[width=\columnwidth]{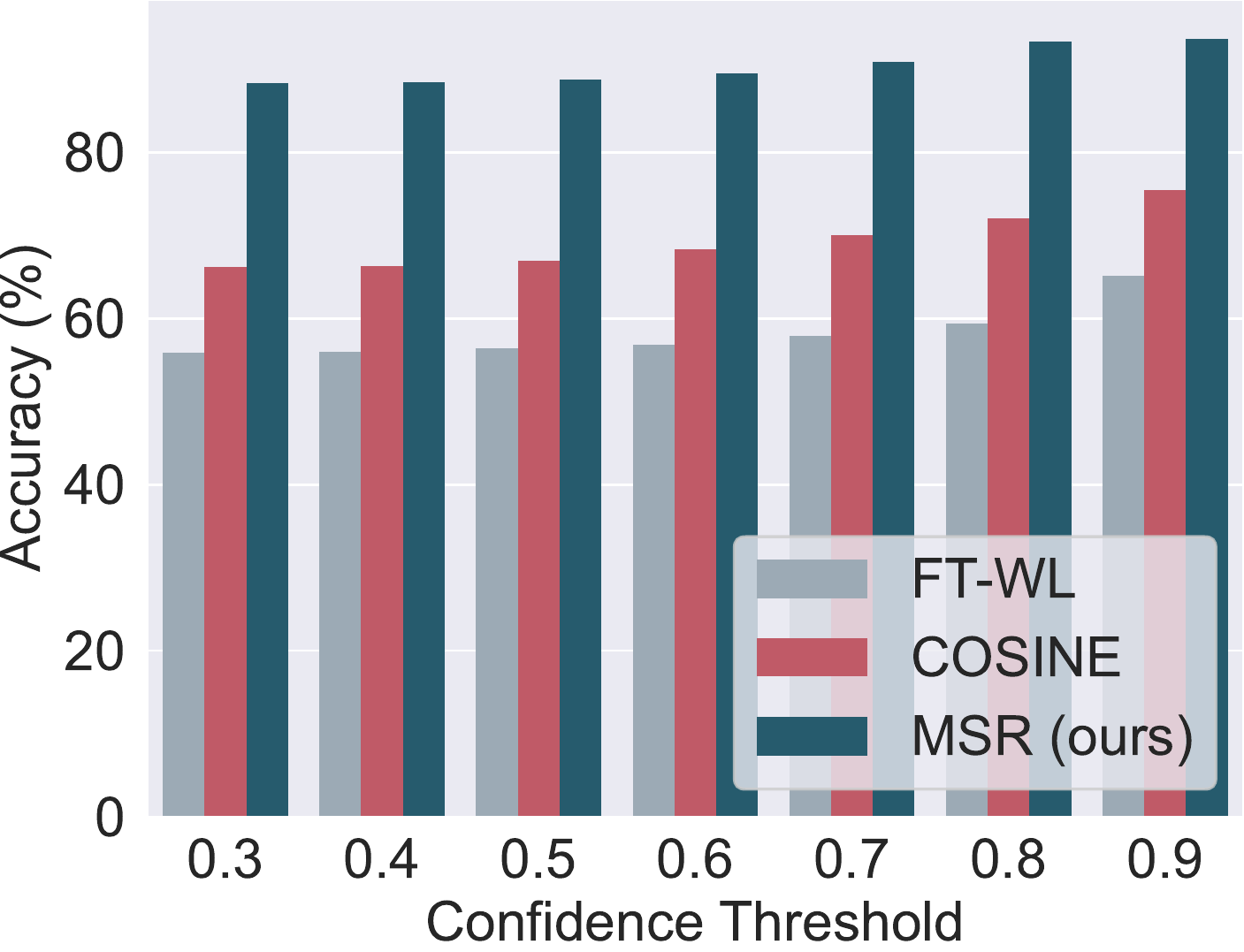}
         \caption{TREC}
    \end{subfigure}\hfill
     \begin{subfigure}[t]{0.5\columnwidth}
         \centering
         \includegraphics[width=\columnwidth]{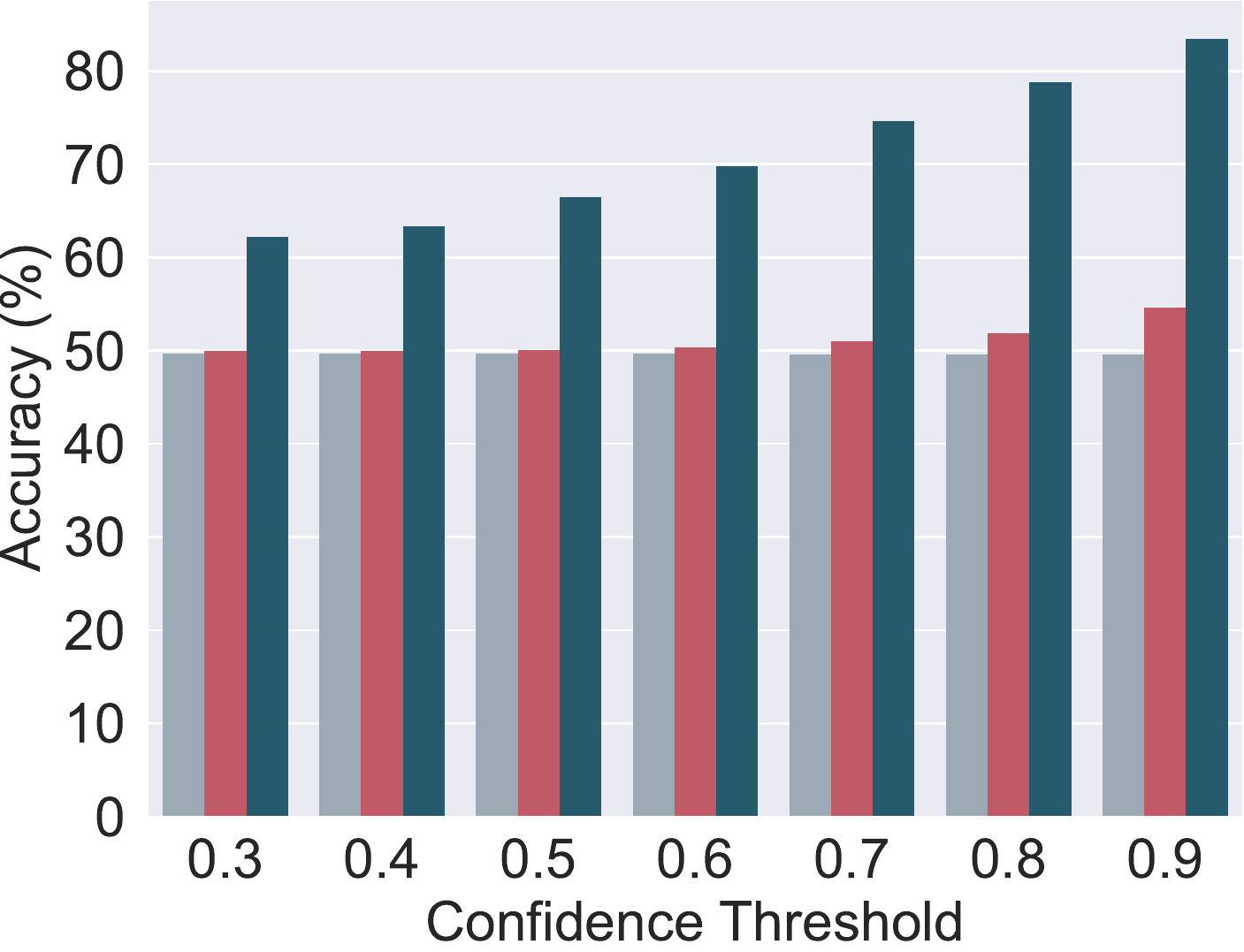}
         \caption{Hausa}
    \end{subfigure}
    \caption{Accuracy \textit{vs.} confidence thresholds.}
    \label{fig:conf_acc_bar}
\end{figure}

\paragraph{Error Decomposition.}
\label{sec:robustness}
Let $y^\prime,\hat{y},y$ denote the model prediction, the noisy weak label, and the clean label, respectively. To investigate how the label noise influences the model predictions, we decompose model prediction errors into three types: (1) Type-A error: $y^\prime=\hat{y};\hat{y}\neq y$ (2) Type-B error: $y^\prime \neq \hat{y} \neq y$ and (3) Type-C error: $y^\prime \neq y;y = \hat{y}$. Type-A/B errors correspond to situations in which a model complies with an incorrect weak label $\hat{y}$, or predicts another incorrect class label. If, on the other hand, the weak label $\hat{y}$ is correct, a Type-C error arises if the model predicts a label different than $\hat{y}$. A higher Type-A error rate indicates that a model memorizes more label noise from the weak sources, while a model that underfits fails to learn useful knowledge from the weak sources can have a higher Type-C error rate.

Figure \ref{fig:robustness_plot} visualizes the three types of errors on three challenging datasets: TREC, Hausa and CoNLL-03. The blue bars represent model robustness, i.e., how often the model predicts correctly when $\hat{y} \neq y$. It clearly shows that direct fine-tuning on weak labels (FT-WL) has a much higher Type-A error rate compared with the model trained on clean data (FT-CL), suggesting that the model quickly memorizes the label noise. On the other hand, the disparity in type C error rate is much smaller, indicating that all models do not underfit and the knowledge from the weak sources is properly transferred. The Type-B error shows similar trends and does not differ much across models. Overall, Type-A error has the strongest impact on model performance. \emph{All the noisy-handling models mainly help with reducing Type-A errors}. We also observe that while COSINE reduces Type-A errors on TREC, it barely works on the other two datasets. Only MSR manages to consistently reduce Type-A errors by over 20\% on all three datasets.

\begin{figure*}[ht!]
        \centering
        
     \begin{subfigure}[b]{0.67\columnwidth}
         \centering
         \includegraphics[width=\columnwidth]{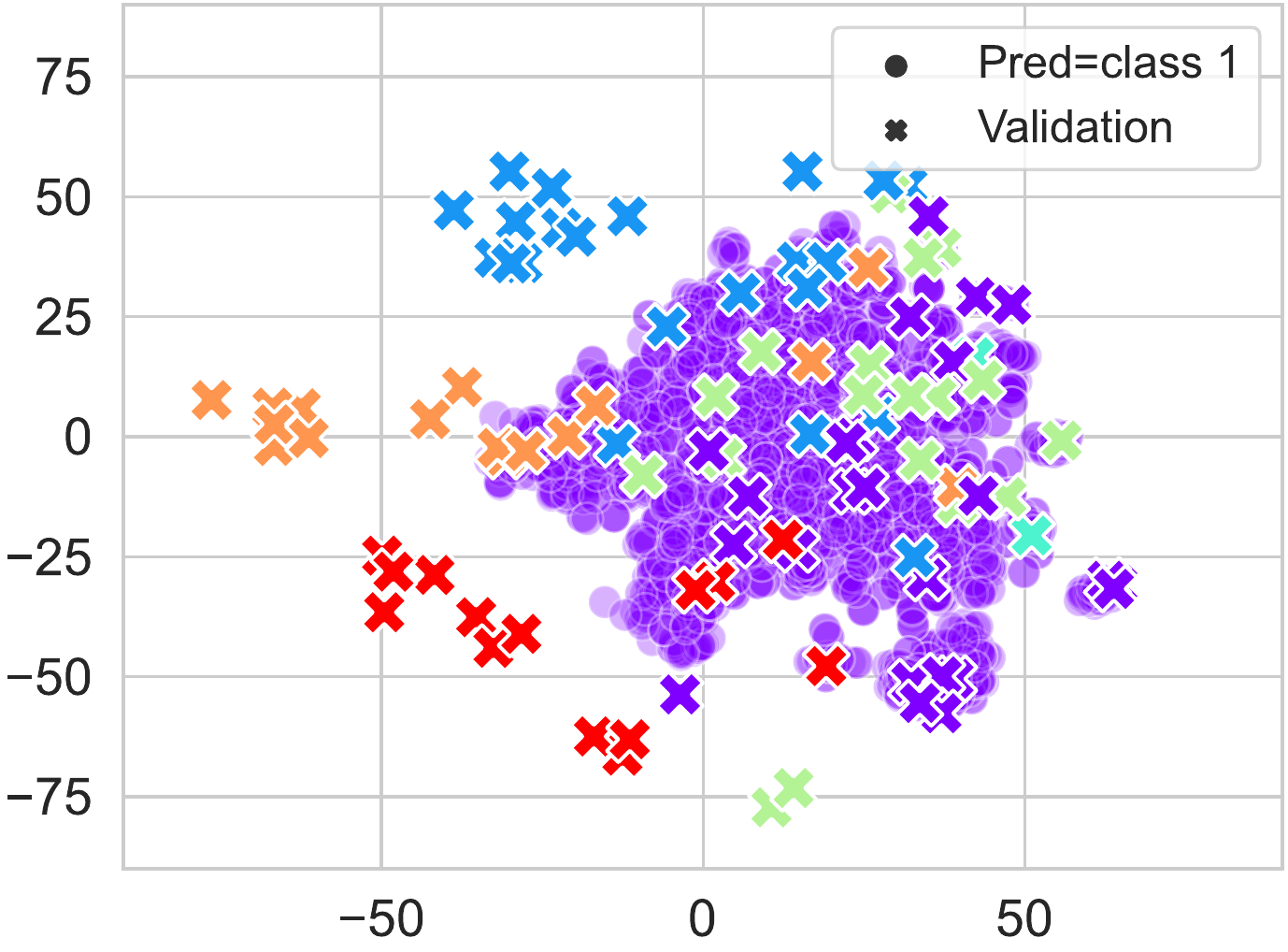}
         \caption{Teacher-Init}
         \label{subfig:val_above_train_feature_bert_wl}
     \end{subfigure}\hfill
      \begin{subfigure}[b]{0.67\columnwidth}
         \centering
         \includegraphics[width=\columnwidth]{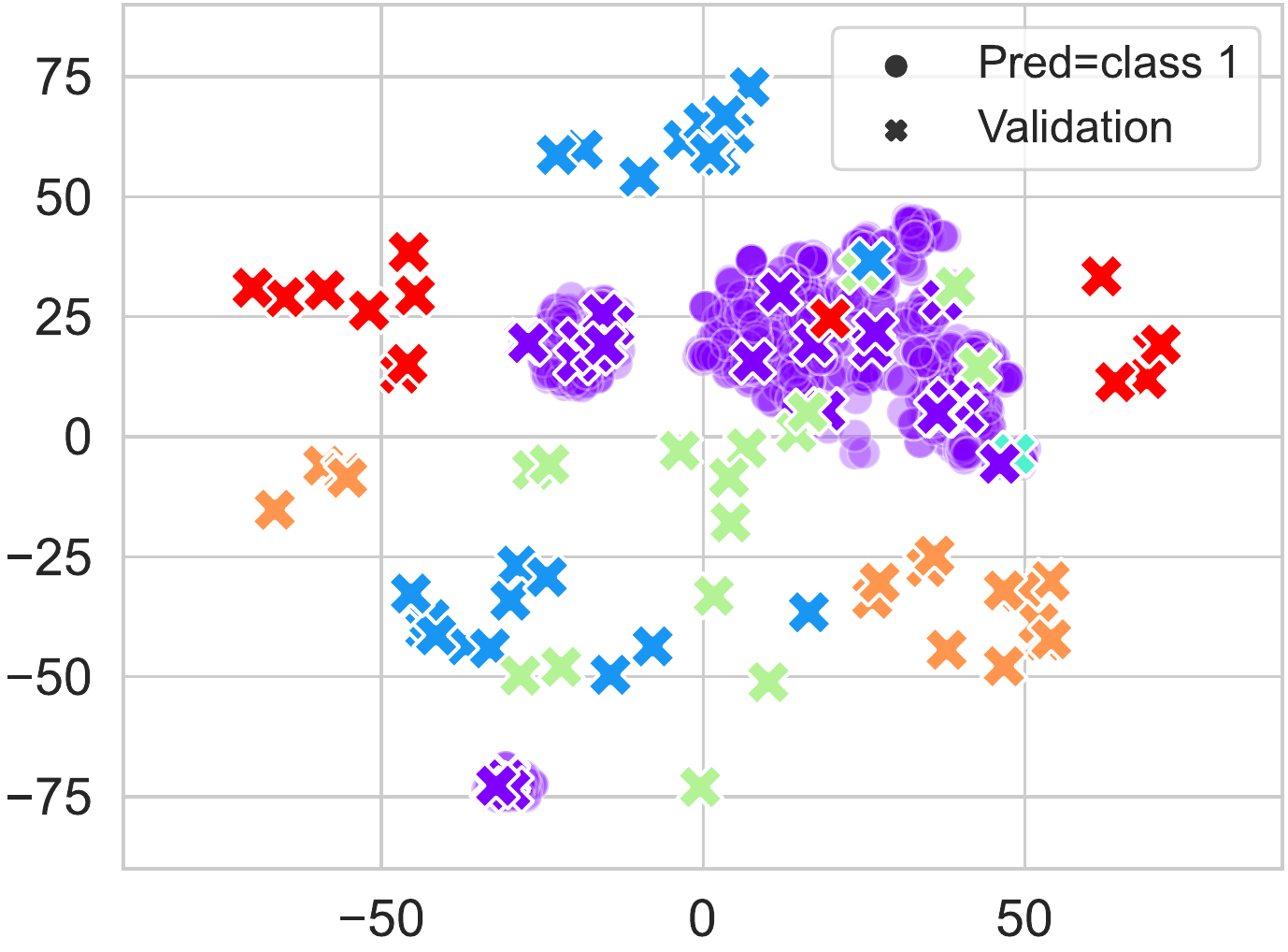}
         \caption{MSR student at step 1500}
         \label{subfig:val_above_train_feature_cosine}
     \end{subfigure}\hfill
     \begin{subfigure}[b]{0.67\columnwidth}
         \centering
         \includegraphics[width=\columnwidth]{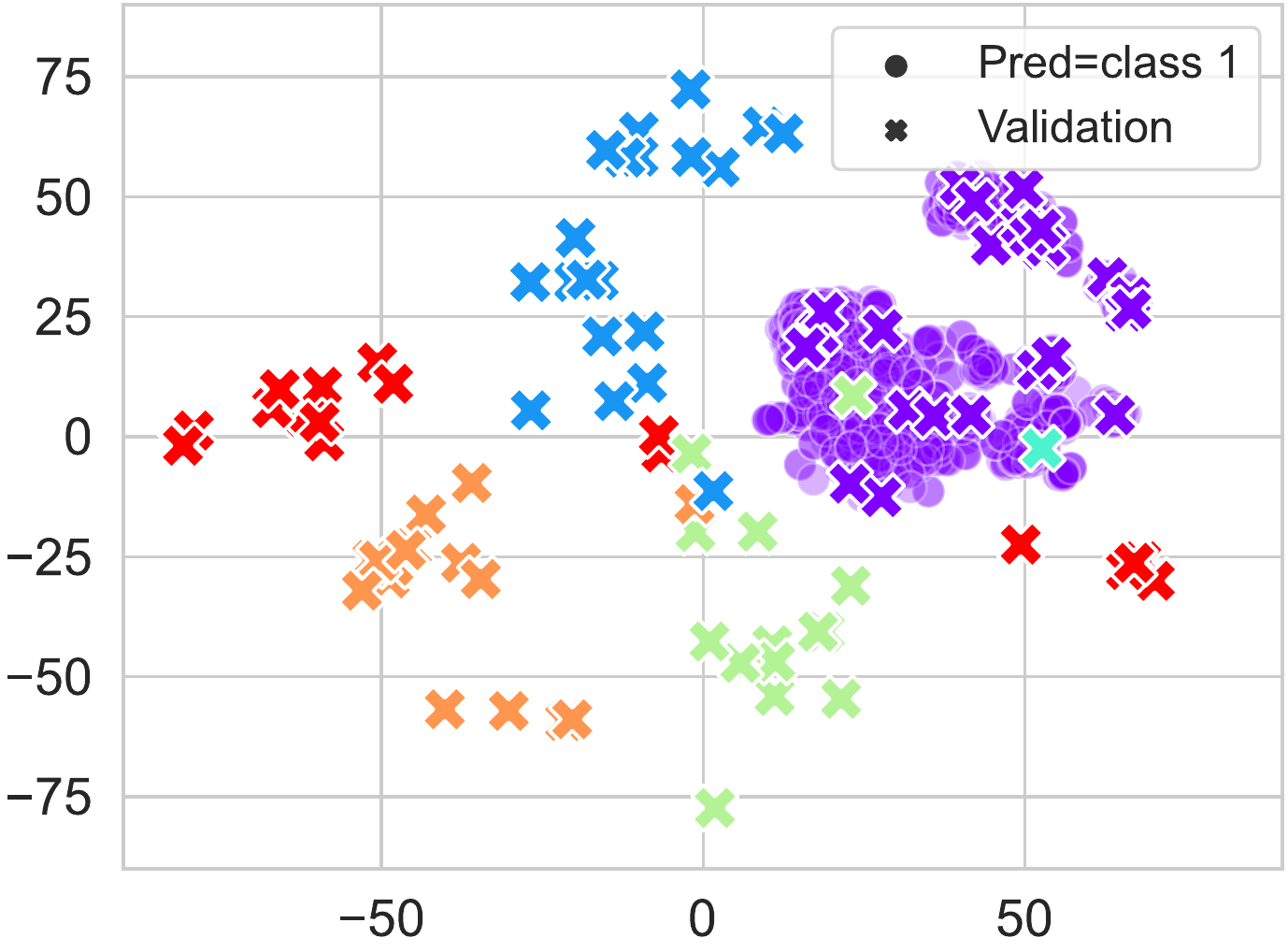}
         \caption{MSR student at step 3000}
         \label{subfig:val_above_train_feature_msr}
     \end{subfigure}\hfill
        
     \begin{subfigure}[b]{0.67\columnwidth}
         \centering
         \includegraphics[width=\columnwidth]{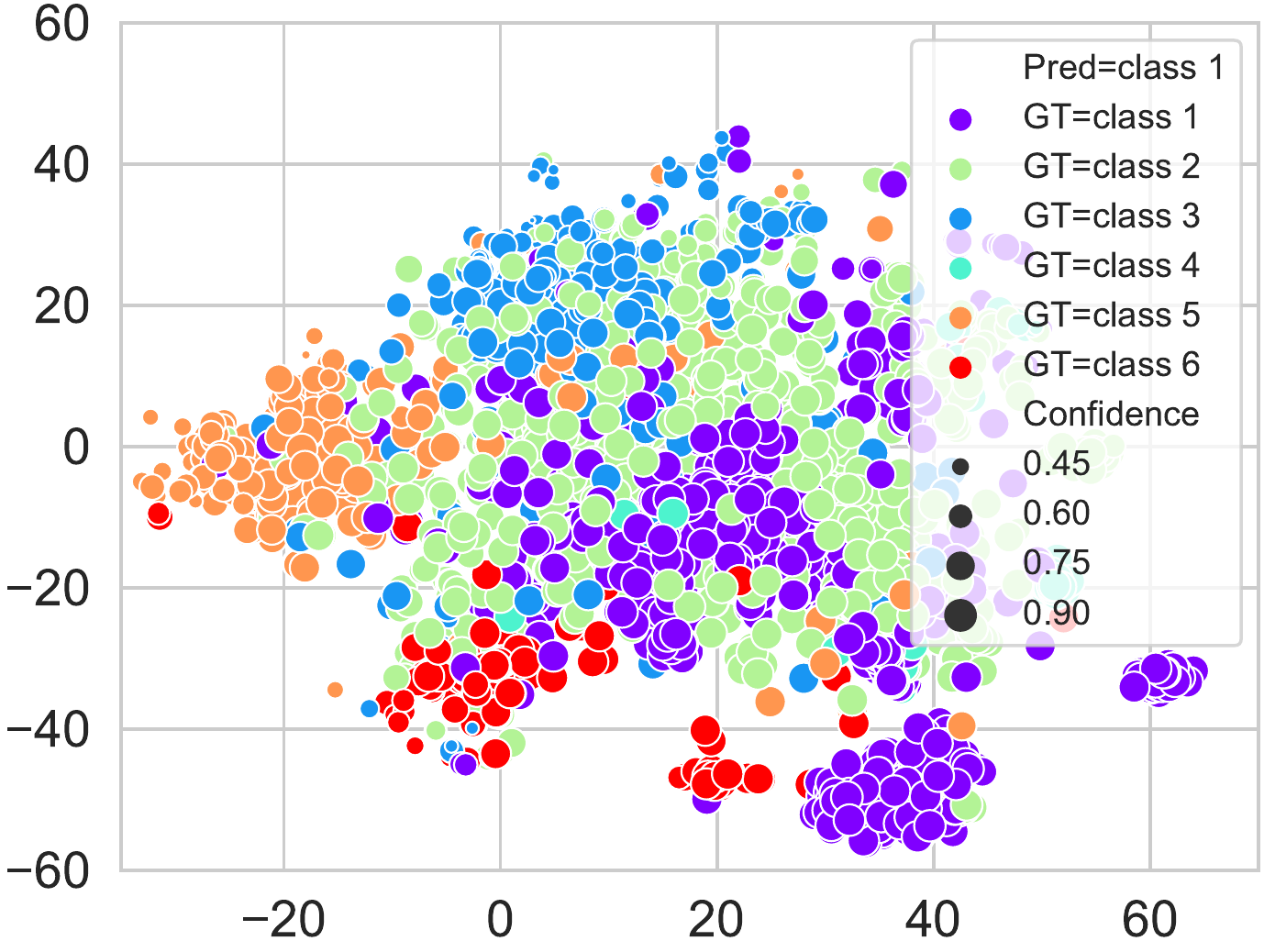}
         \caption{Teacher-Init}
         \label{subfig:feature_bert_wl}
     \end{subfigure}\hfill
      \begin{subfigure}[b]{0.67\columnwidth}
         \centering
         \includegraphics[width=\columnwidth]{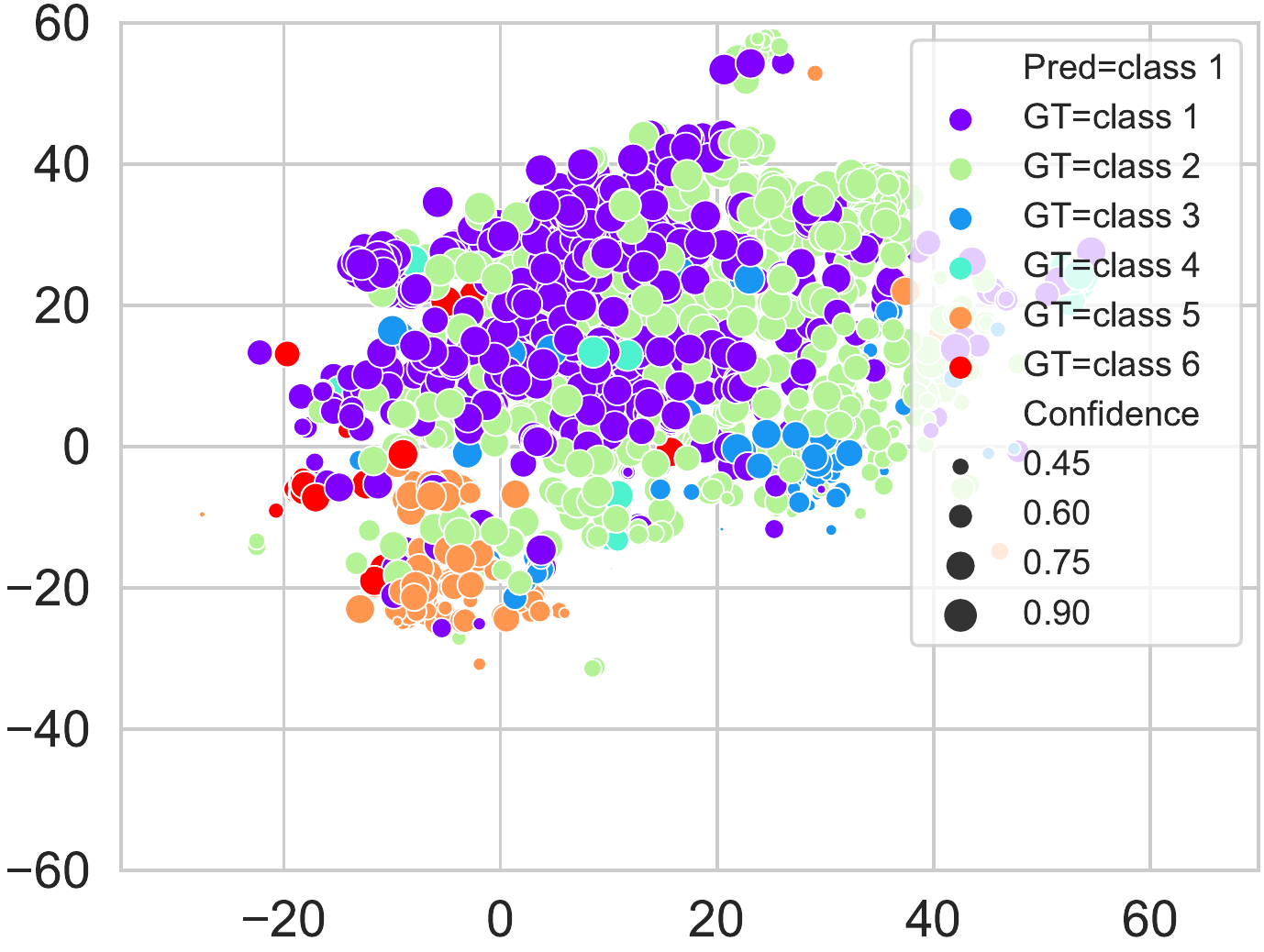}
         \caption{COSINE}
         \label{subfig:feature_cosine}
     \end{subfigure}\hfill
     \begin{subfigure}[b]{0.67\columnwidth}
         \centering
         \includegraphics[width=\columnwidth]{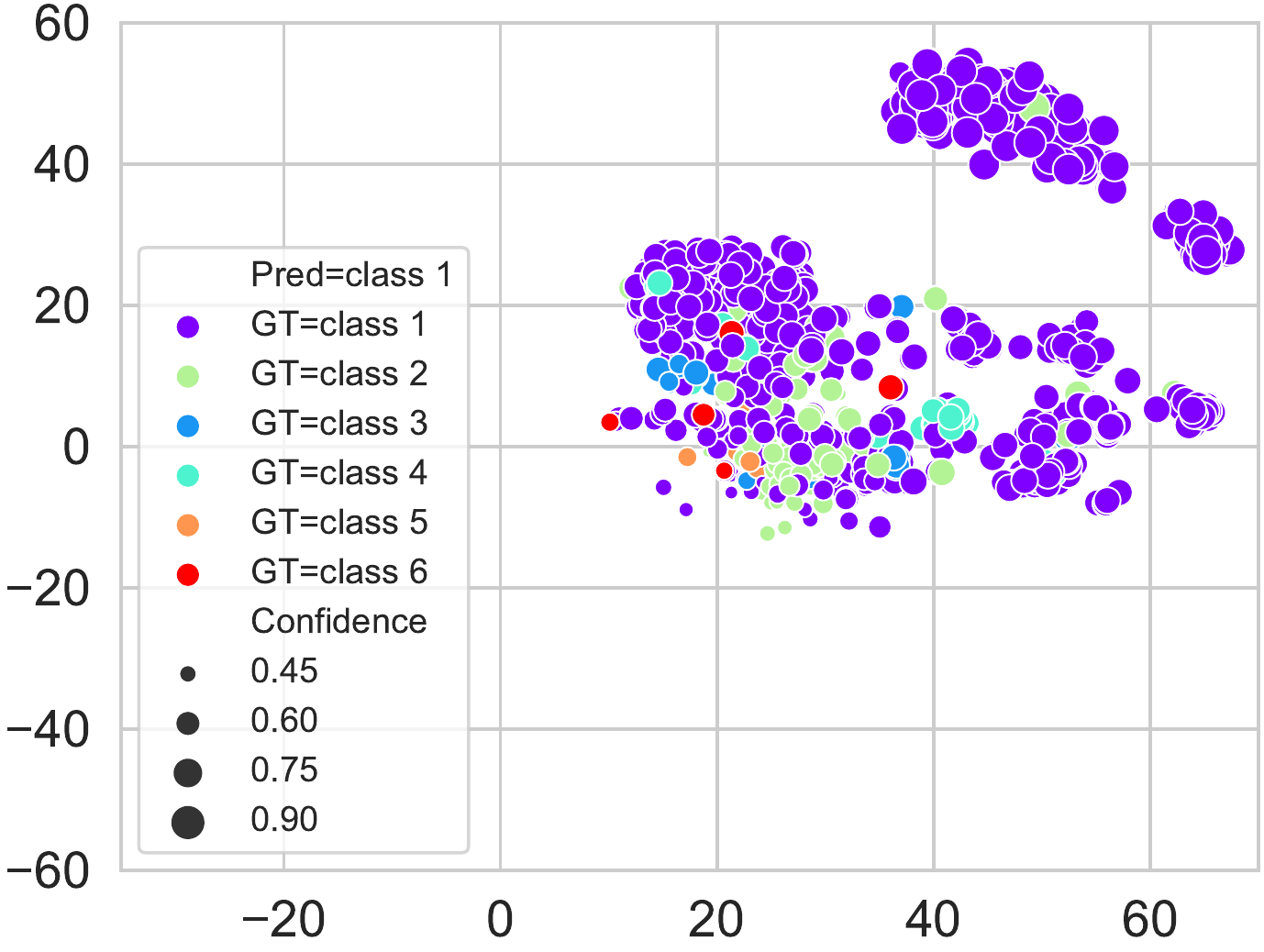}
         \caption{MSR}
         \label{subfig:feature_msr}
     \end{subfigure}\hfill
        
        \caption{Projected feature space of different models on TREC using t-SNE {\cite{van2008visualizing_tsne}}. The circles represent training samples that are predicted as class 1. \textbf{a)-c)}: development of MSR during training. Circles are colored by the predicted class (i.e., class 1, in purple). The validation samples are represented by crosses and colored according to the ground truth labels. The MSR student gradually improves its feature space to embed the training and validation samples from the same class in the same area. \textbf{d)-f)}: training samples are colored according to their ground truth labels; model confidence is reflected by the size of the circles. Teacher-Init and COSINE misclassify samples with high confidence. MSR attains a cleaner cluster.}
        \label{fig:tsne}
\end{figure*}

\paragraph{Accuracy vs Confidence.}
As confidence-based filtering is a key component in both COSINE and MSR, we show the accuracy of model predictions with different confidence thresholds in Figure~\ref{fig:conf_acc_bar}. As can be seen, \emph{even using a high confidence threshold for COSINE, the accuracy is still low}, which is why it struggles to improve on challenging datasets. MSR, on the contrary, consistently attains higher accuracy with higher confidence thresholds, and thereby confidence-based filtering on top of MSR help lead to better performance.

\paragraph{Impact of Label Noise on Feature Space.}
We also analyze how the label noise influences representation learning. Figure \ref{fig:tsne} illustrates the projected feature space of different models on TREC. For a clear visualization, we present only training samples predicted as class 1 by the models in the form of circles. In~\cref{subfig:val_above_train_feature_bert_wl,subfig:val_above_train_feature_cosine,subfig:val_above_train_feature_msr}, we further visualize the feature space of validation samples (represented by crosses). As can be seen, initially the feature space of class 1 overlaps with that of other classes from the validation set. As the training proceeds, when the teacher keeps refining itself, the MSR student gradually reduces such overlap and learns a well-split representation space. In~\cref{subfig:feature_bert_wl,subfig:feature_cosine,subfig:feature_msr}, we compare the feature space between different models. The training samples are colored according to their ground truth classes to highlight the misclassification ratio (the more colorful the clusters, the higher the misclassification ratio). We observe that Teacher-Init makes many wrong predictions with high degree of confidence. In this case, utilizing the confidence score for denoising is fragile. This may explain why COSINE, despite offering a more compact cluster, still has a considerable amount of misclassification. Finally, MSR has a considerably cleaner cluster and is less affected by error propagation than COSINE.

\begin{figure}[t]
    \centering
     \begin{subfigure}[t]{0.5\columnwidth}
         \centering
         \includegraphics[width=\columnwidth]{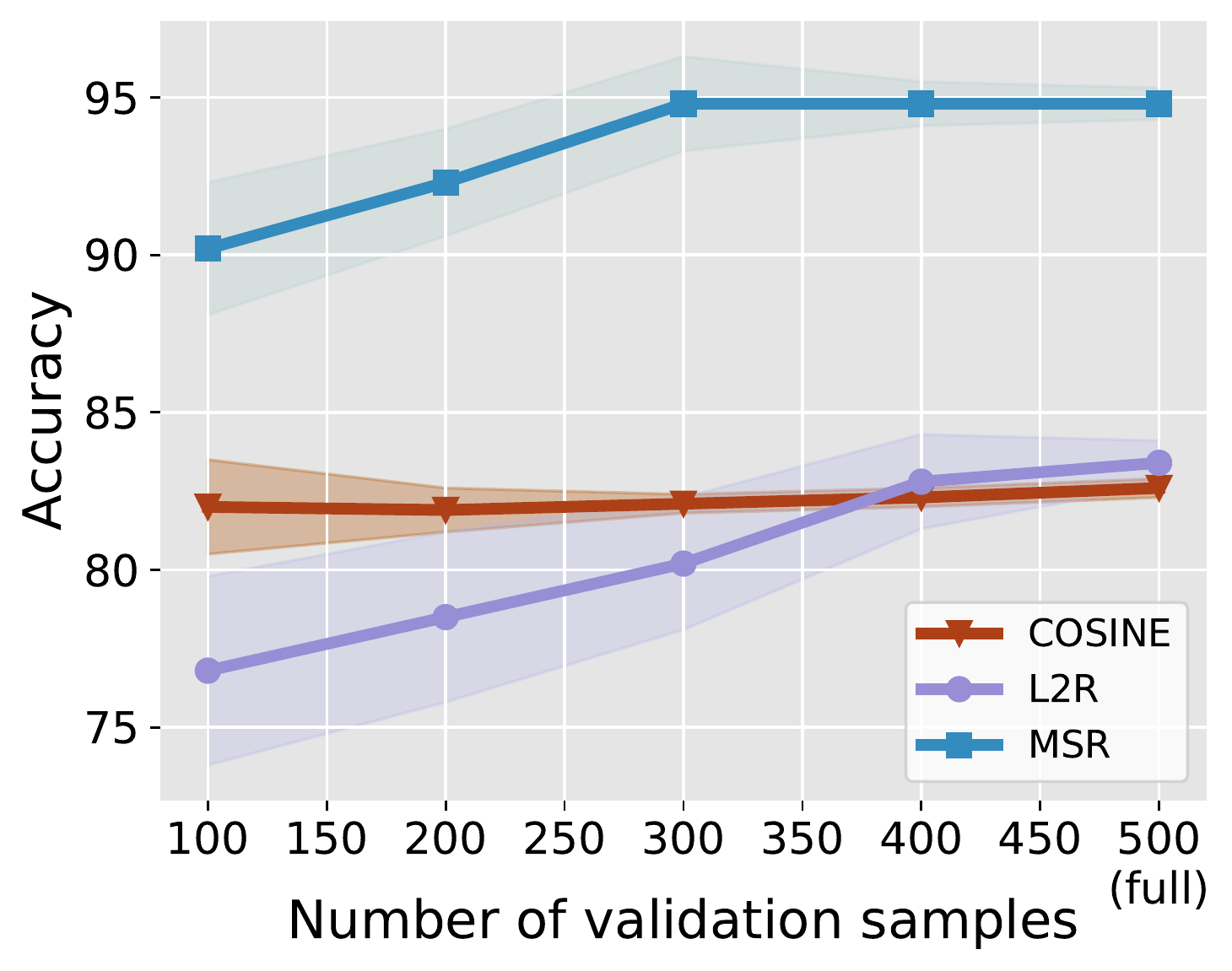}
         \caption{TREC}
    \end{subfigure}\hfill
     \begin{subfigure}[t]{0.5\columnwidth}
         \centering
         \includegraphics[width=\columnwidth]{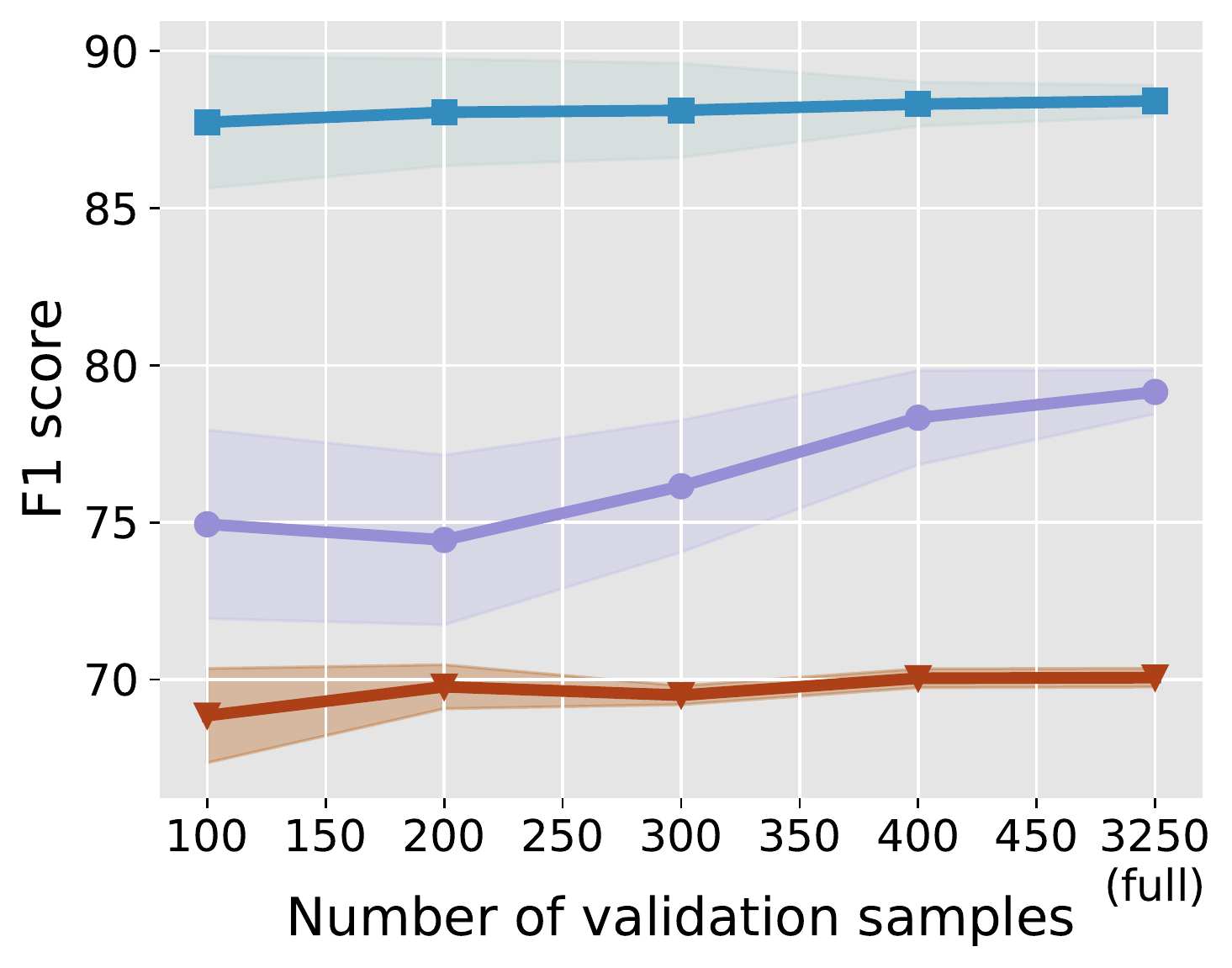}
         \caption{CoNLL-03}
    \end{subfigure}
    \caption{Accuracy \textit{vs.} number of validation samples.}
    \label{fig:val_size}
\end{figure}

\paragraph{Effects of Validation Data Size.}
The model performance reported in Table~\ref{tab:main_result} is based on the original data splits from the WRENCH benchmark. The size of the validation sets is mostly less than 15\% of the training sets. Typically, they are used to perform early stopping and model selection. For meta-learning based methods, they additionally rely on the validation sets for meta-update and might be more sensitive to validation size. Hence, we study how the validation size affects different models. In particular, we randomly sample a subset from the original validation set $\mathcal{D}_{v}$ and repeat the same training process. Figure \ref{fig:val_size} presents the results. We find that the validation size indeed has a greater impact on meta-learning approaches. However, \emph{MSR still retains its high generalization performance even with as few as 100 validation samples}, suggesting that MSR is very data efficient in performing the self-refinement.

\begin{table}[h!]
    \centering
    \resizebox{\columnwidth}{!}{
    \begin{tabular}{ccc}
    \toprule  
    \textbf{Configuration}     &  \begin{tabular}[c]{@{}c@{}} \textbf{Seq. Classification}\\ \textbf{(Acc)} \end{tabular} &  \begin{tabular}[c]{@{}c@{}} \textbf{NER}\\ \textbf{(F1)} \end{tabular}   \\\midrule   
    Teacher-Init   &  73.75 &  68.99 \\\midrule   
    Student   &  \textbf{83.43} &  \textbf{81.50}\\
    Teacher  &  82.38 ($\downarrow$ 1.05\%) &  80.26 ($\downarrow$ 1.24\%)\\
  w\slash o Teacher Scheduler  &  81.80 ($\downarrow$ 1.63\%)  &  80.15 ($\downarrow$ 1.35\%)\\
  w\slash o Confidence Filtering &  82.32 ($\downarrow$ 1.11\%)  &  81.09 ($\downarrow$ 0.41\%)\\
  w\slash o Both &  81.63 ($\downarrow$ 1.80\%)  &  79.95 ($\downarrow$ 1.55\%)\\
    \bottomrule
    \end{tabular}
    }
    \caption{Summary of ablation experiments aggregated across multiple datasets. See Appendix \ref{sec:ablation_studies_detailed} for results in each dataset.}
    \label{tab:ablation_results_summary}
\end{table}

\paragraph{Ablation Study.}
\label{sec: ablation_study}
Table \ref{tab:ablation_results_summary} summarizes the impact of different components of our method. In general, our student model performs slightly better than the teacher. This is as expected because a) the teacher's goal is to guide the student to generalize better, the training loss does not explicitly encourage the teacher to improve its accuracy, and b) the confidence filtering helps the student avoid fitting some wrong pseudo-labels from the teacher. This is also confirmed by the decreased performance when the filter is removed. In addition, applying a learning rate scheduler is better than using a fixed learning rate throughout training.

\section{Conclusion}
We present MSR, a meta-learning based self-refinement framework that enables robust learning with weak labels. Unlike conventional self-training which relies on a fixed teacher, MSR dynamically refines the teacher based on the student's performance on the validation set. To further suppress error propagation, we introduce a learning rate scheduler to the teacher and add confidence filtering to the student. We demonstrate that our framework performs on par with or better than current SOTAs on both sequence classification and labeling tasks.

\section*{Limitations}
In this work, Our primary focus is to propose a strong weak supervision method that works reliably under various weak supervision settings. We employ meta-learning techniques to address the issue of unreliable confidence scores under challenging settings (Figure \ref{fig:conf_acc_bar}). Despite the effectiveness, the main limitation of our method, just like other meta-learning based frameworks, is the computational overhead. The teacher update step (Algorithm \ref{algorithm}, Line 4-6) requires computing both the first and second-order derivatives, which incurs additional computation time and higher memory consumption. Consequently, our method requires longer training.\footnote{Detailed training time on each dataset can be found in Appendix \ref{appendix:averge_runtime_msr} The most costly training of MSR takes roughly 3 hours.} Implementation-wise and computation-wise, MSR is as complex as other existing meta-learning based methods, like L2R \cite{l2r_18} and MW-Net \cite{meta_weight19}, but performs substantially better than them in all weak supervision scenarios we evaluated. It is worth noting that MSR has \emph{no overhead at inference time}. In weak supervision, the data annotation cost is considered the most significant bottleneck. A stronger model is often obtained by trading some more computation with the cost and effort of obtaining more human-generated, manual annotations. Hence, the one-off investment of training MSR can be worthwhile for real-world weak supervision applications.

\section*{Acknowledgments}
This work has been partially funded by the Deutsche Forschungsgemeinschaft (DFG, German Research Foundation) – Project-ID 232722074 – SFB 1102 and the EU Horizon 2020 projects ROXANNE under grant number 833635.

\bibliography{eacl23}

\begin{thebibliography}{43}
\expandafter\ifx\csname natexlab\endcsname\relax\def\natexlab#1{#1}\fi

\bibitem[{Angluin and Laird(1988)}]{angluin1988learning}
Dana Angluin and Philip Laird. 1988.
\newblock Learning from noisy examples.
\newblock \emph{Machine Learning}, 2(4):343--370.

\bibitem[{Awasthi et~al.(2020)Awasthi, Ghosh, Goyal, and
  Sarawagi}]{implyloss_2020}
Abhijeet Awasthi, Sabyasachi Ghosh, Rasna Goyal, and Sunita Sarawagi. 2020.
\newblock \href {https://openreview.net/forum?id=SkeuexBtDr} {Learning from
  rules generalizing labeled exemplars}.
\newblock In \emph{8th International Conference on Learning Representations,
  {ICLR} 2020, Addis Ababa, Ethiopia, April 26-30, 2020}. OpenReview.net.

\bibitem[{Bekker and Goldberger(2016)}]{DBLP:conf/icassp/BekkerG16}
Alan~Joseph Bekker and Jacob Goldberger. 2016.
\newblock \href {https://doi.org/10.1109/ICASSP.2016.7472164} {Training deep
  neural-networks based on unreliable labels}.
\newblock In \emph{2016 {IEEE} International Conference on Acoustics, Speech
  and Signal Processing, {ICASSP} 2016, Shanghai, China, March 20-25, 2016},
  pages 2682--2686. {IEEE}.

\bibitem[{Davis et~al.(2013)Davis, Wiegers, Roberts, King, Lay,
  Lennon{-}Hopkins, Sciaky, Johnson, Keating, Greene, Hernandez, McConnell,
  Enayetallah, and Mattingly}]{DBLP:journals/biodb/DavisWRKLLSJKGHMEM13}
Allan~Peter Davis, Thomas~C. Wiegers, Phoebe~M. Roberts, Benjamin~L. King,
  Jean~M. Lay, Kelley Lennon{-}Hopkins, Daniela Sciaky, Robin~J. Johnson,
  Heather Keating, Nigel Greene, Robert Hernandez, Kevin~J. McConnell, Ahmed
  Enayetallah, and Carolyn~J. Mattingly. 2013.
\newblock \href {https://doi.org/10.1093/database/bat080} {A ctd-pfizer
  collaboration: manual curation of 88 000 scientific articles text mined for
  drug-disease and drug-phenotype interactions}.
\newblock \emph{Database J. Biol. Databases Curation}, 2013.

\bibitem[{Devlin et~al.(2019)Devlin, Chang, Lee, and Toutanova}]{devlinCLT19}
Jacob Devlin, Ming{-}Wei Chang, Kenton Lee, and Kristina Toutanova. 2019.
\newblock \href {https://doi.org/10.18653/v1/n19-1423} {{BERT:} pre-training of
  deep bidirectional transformers for language understanding}.
\newblock In \emph{Proceedings of the 2019 Conference of the North American
  Chapter of the Association for Computational Linguistics: Human Language
  Technologies, {NAACL-HLT} 2019, Minneapolis, MN, USA, June 2-7, 2019, Volume
  1 (Long and Short Papers)}, pages 4171--4186. Association for Computational
  Linguistics.

\bibitem[{Fu et~al.(2020)Fu, Chen, Sala, Hooper, Fatahalian, and
  R{\'e}}]{fu2020fast_flying_squid}
Daniel Fu, Mayee Chen, Frederic Sala, Sarah Hooper, Kayvon Fatahalian, and
  Christopher R{\'e}. 2020.
\newblock Fast and three-rious: Speeding up weak supervision with triplet
  methods.
\newblock In \emph{International Conference on Machine Learning}, pages
  3280--3291. PMLR.

\bibitem[{Gal and Ghahramani(2016)}]{mc_drop_uncertain_16}
Yarin Gal and Zoubin Ghahramani. 2016.
\newblock \href {http://proceedings.mlr.press/v48/gal16.html} {Dropout as a
  bayesian approximation: Representing model uncertainty in deep learning}.
\newblock In \emph{Proceedings of the 33nd International Conference on Machine
  Learning, {ICML} 2016, New York City, NY, USA, June 19-24, 2016}, volume~48
  of \emph{{JMLR} Workshop and Conference Proceedings}, pages 1050--1059.
  JMLR.org.

\bibitem[{Goldberger and Ben{-}Reuven(2017)}]{goldbergerB17}
Jacob Goldberger and Ehud Ben{-}Reuven. 2017.
\newblock \href {https://openreview.net/forum?id=H12GRgcxg} {Training deep
  neural-networks using a noise adaptation layer}.
\newblock In \emph{5th International Conference on Learning Representations,
  {ICLR} 2017, Toulon, France, April 24-26, 2017, Conference Track
  Proceedings}. OpenReview.net.

\bibitem[{Gu et~al.(2021)Gu, Masotto, Bachani, Lakshminarayanan, Nikodem, and
  Yin}]{gu2021instance}
Keren Gu, Xander Masotto, Vandana Bachani, Balaji Lakshminarayanan, Jack
  Nikodem, and Dong Yin. 2021.
\newblock An instance-dependent simulation framework for learning with label
  noise.
\newblock \emph{arXiv preprint arXiv:2107.11413}.

\bibitem[{Han et~al.(2018)Han, Yao, Yu, Niu, Xu, Hu, Tsang, and
  Sugiyama}]{DBLP:conf/nips/HanYYNXHTS18}
Bo~Han, Quanming Yao, Xingrui Yu, Gang Niu, Miao Xu, Weihua Hu, Ivor~W. Tsang,
  and Masashi Sugiyama. 2018.
\newblock \href
  {https://proceedings.neurips.cc/paper/2018/hash/a19744e268754fb0148b017647355b7b-Abstract.html}
  {Co-teaching: Robust training of deep neural networks with extremely noisy
  labels}.
\newblock In \emph{Advances in Neural Information Processing Systems 31: Annual
  Conference on Neural Information Processing Systems 2018, NeurIPS 2018,
  December 3-8, 2018, Montr{\'{e}}al, Canada}, pages 8536--8546.

\bibitem[{Hedderich et~al.(2020)Hedderich, Adelani, Zhu, Alabi, Markus, and
  Klakow}]{DBLP:conf/emnlp/HedderichAZAMK20}
Michael~A. Hedderich, David~Ifeoluwa Adelani, Dawei Zhu, Jesujoba~O. Alabi,
  Udia Markus, and Dietrich Klakow. 2020.
\newblock \href {https://doi.org/10.18653/v1/2020.emnlp-main.204} {Transfer
  learning and distant supervision for multilingual transformer models: {A}
  study on african languages}.
\newblock In \emph{Proceedings of the 2020 Conference on Empirical Methods in
  Natural Language Processing, {EMNLP} 2020, Online, November 16-20, 2020},
  pages 2580--2591. Association for Computational Linguistics.

\bibitem[{Hendrycks et~al.(2018)Hendrycks, Mazeika, Wilson, and
  Gimpel}]{DBLP:conf/nips/HendrycksMWG18}
Dan Hendrycks, Mantas Mazeika, Duncan Wilson, and Kevin Gimpel. 2018.
\newblock \href
  {https://proceedings.neurips.cc/paper/2018/hash/ad554d8c3b06d6b97ee76a2448bd7913-Abstract.html}
  {Using trusted data to train deep networks on labels corrupted by severe
  noise}.
\newblock In \emph{Advances in Neural Information Processing Systems 31: Annual
  Conference on Neural Information Processing Systems 2018, NeurIPS 2018,
  December 3-8, 2018, Montr{\'{e}}al, Canada}, pages 10477--10486.

\bibitem[{Karamanolakis et~al.(2021)Karamanolakis, Mukherjee, Zheng, and
  Awadallah}]{astra_karamanolakis2021}
Giannis Karamanolakis, Subhabrata~(Subho) Mukherjee, Guoqing Zheng, and
  Ahmed~H. Awadallah. 2021.
\newblock \href
  {https://www.microsoft.com/en-us/research/publication/self-training-weak-supervision-astra/}
  {Self-training with weak supervision}.
\newblock In \emph{NAACL 2021}. NAACL 2021.

\bibitem[{Lee et~al.(2013)}]{self-training_lee2013pseudo}
Dong-Hyun Lee et~al. 2013.
\newblock Pseudo-label: The simple and efficient semi-supervised learning
  method for deep neural networks.
\newblock In \emph{Workshop on challenges in representation learning, ICML},
  volume~3, page 896.

\bibitem[{Li et~al.(2020)Li, Socher, and Hoi}]{liSH20}
Junnan Li, Richard Socher, and Steven C.~H. Hoi. 2020.
\newblock \href {https://openreview.net/forum?id=HJgExaVtwr} {Dividemix:
  Learning with noisy labels as semi-supervised learning}.
\newblock In \emph{8th International Conference on Learning Representations,
  {ICLR} 2020, Addis Ababa, Ethiopia, April 26-30, 2020}. OpenReview.net.

\bibitem[{Li and Roth(2002)}]{trec_li2002learning}
Xin Li and Dan Roth. 2002.
\newblock Learning question classifiers.
\newblock In \emph{COLING 2002: The 19th International Conference on
  Computational Linguistics}.

\bibitem[{Liang et~al.(2020)Liang, Yu, Jiang, Er, Wang, Zhao, and
  Zhang}]{liang2020bond}
Chen Liang, Yue Yu, Haoming Jiang, Siawpeng Er, Ruijia Wang, Tuo Zhao, and Chao
  Zhang. 2020.
\newblock Bond: Bert-assisted open-domain named entity recognition with distant
  supervision.
\newblock In \emph{ACM SIGKDD International Conference on Knowledge Discovery
  and Data Mining}.

\bibitem[{Lison et~al.(2020)Lison, Barnes, Hubin, and Touileb}]{lison2020_ner}
Pierre Lison, Jeremy Barnes, Aliaksandr Hubin, and Samia Touileb. 2020.
\newblock \href {https://doi.org/10.18653/v1/2020.acl-main.139} {Named entity
  recognition without labelled data: {A} weak supervision approach}.
\newblock In \emph{Proceedings of the 58th Annual Meeting of the Association
  for Computational Linguistics, {ACL} 2020, Online, July 5-10, 2020}, pages
  1518--1533. Association for Computational Linguistics.

\bibitem[{Liu et~al.(2019)Liu, Ott, Goyal, Du, Joshi, Chen, Levy, Lewis,
  Zettlemoyer, and Stoyanov}]{liu19roberta}
Yinhan Liu, Myle Ott, Naman Goyal, Jingfei Du, Mandar Joshi, Danqi Chen, Omer
  Levy, Mike Lewis, Luke Zettlemoyer, and Veselin Stoyanov. 2019.
\newblock \href {http://arxiv.org/abs/1907.11692} {Roberta: {A} robustly
  optimized {BERT} pretraining approach}.
\newblock \emph{CoRR}, abs/1907.11692.

\bibitem[{Loshchilov and Hutter(2019)}]{LoshchilovH19_adamW}
Ilya Loshchilov and Frank Hutter. 2019.
\newblock \href {https://openreview.net/forum?id=Bkg6RiCqY7} {Decoupled weight
  decay regularization}.
\newblock In \emph{7th International Conference on Learning Representations,
  {ICLR} 2019, New Orleans, LA, USA, May 6-9, 2019}. OpenReview.net.

\bibitem[{Maas et~al.(2011)Maas, Daly, Pham, Huang, Ng, and
  Potts}]{imdb_maas-etal-2011-learning}
Andrew~L. Maas, Raymond~E. Daly, Peter~T. Pham, Dan Huang, Andrew~Y. Ng, and
  Christopher Potts. 2011.
\newblock \href {https://aclanthology.org/P11-1015} {Learning word vectors for
  sentiment analysis}.
\newblock In \emph{Proceedings of the 49th Annual Meeting of the Association
  for Computational Linguistics: Human Language Technologies}, pages 142--150,
  Portland, Oregon, USA. Association for Computational Linguistics.

\bibitem[{Menon et~al.(2016)Menon, van Rooyen, and
  Natarajan}]{DBLP:journals/corr/MenonRN16}
Aditya~Krishna Menon, Brendan van Rooyen, and Nagarajan Natarajan. 2016.
\newblock \href {http://arxiv.org/abs/1605.00751} {Learning from binary labels
  with instance-dependent corruption}.
\newblock \emph{CoRR}, abs/1605.00751.

\bibitem[{Mukherjee and Hassan~Awadallah(2020)}]{mukherjee-awadallah-2020-ust}
Subhabrata Mukherjee and Ahmed Hassan~Awadallah. 2020.
\newblock \href
  {https://papers.nips.cc/paper/2020/file/f23d125da1e29e34c552f448610ff25f-Paper.pdf}
  {Uncertainty-aware self-training for few-shot text classification}.
\newblock In \emph{Advances in Neural Information Processing Systems (NeurIPS
  2020)}, Online.

\bibitem[{Patrini et~al.(2017)Patrini, Rozza, Menon, Nock, and
  Qu}]{patriniRMNQ17}
Giorgio Patrini, Alessandro Rozza, Aditya~Krishna Menon, Richard Nock, and
  Lizhen Qu. 2017.
\newblock \href {https://doi.org/10.1109/CVPR.2017.240} {Making deep neural
  networks robust to label noise: {A} loss correction approach}.
\newblock In \emph{2017 {IEEE} Conference on Computer Vision and Pattern
  Recognition, {CVPR} 2017, Honolulu, HI, USA, July 21-26, 2017}, pages
  2233--2241. {IEEE} Computer Society.

\bibitem[{Pham et~al.(2021)Pham, Dai, Xie, and Le}]{Pham21_meta_pseudo_labels}
Hieu Pham, Zihang Dai, Qizhe Xie, and Quoc~V. Le. 2021.
\newblock \href
  {https://openaccess.thecvf.com/content/CVPR2021/html/Pham\_Meta\_Pseudo\_Labels\_CVPR\_2021\_paper.html}
  {Meta pseudo labels}.
\newblock In \emph{{IEEE} Conference on Computer Vision and Pattern
  Recognition, {CVPR} 2021, virtual, June 19-25, 2021}, pages 11557--11568.
  Computer Vision Foundation / {IEEE}.

\bibitem[{Pradhan et~al.(2013)Pradhan, Moschitti, Xue, Ng, Bj{\"o}rkelund,
  Uryupina, Zhang, and Zhong}]{ontonotes_pradhan-etal-2013-towards}
Sameer Pradhan, Alessandro Moschitti, Nianwen Xue, Hwee~Tou Ng, Anders
  Bj{\"o}rkelund, Olga Uryupina, Yuchen Zhang, and Zhi Zhong. 2013.
\newblock \href {https://aclanthology.org/W13-3516} {Towards robust linguistic
  analysis using {O}nto{N}otes}.
\newblock In \emph{Proceedings of the Seventeenth Conference on Computational
  Natural Language Learning}, pages 143--152, Sofia, Bulgaria. Association for
  Computational Linguistics.

\bibitem[{Ratner et~al.(2017)Ratner, Bach, Ehrenberg, Fries, Wu, and
  R{\'{e}}}]{snorkel_2017}
Alexander Ratner, Stephen~H. Bach, Henry~R. Ehrenberg, Jason~Alan Fries, Sen
  Wu, and Christopher R{\'{e}}. 2017.
\newblock \href {https://doi.org/10.14778/3157794.3157797} {Snorkel: Rapid
  training data creation with weak supervision}.
\newblock \emph{Proc. {VLDB} Endow.}, 11(3):269--282.

\bibitem[{Ren et~al.(2018)Ren, Zeng, Yang, and Urtasun}]{l2r_18}
Mengye Ren, Wenyuan Zeng, Bin Yang, and Raquel Urtasun. 2018.
\newblock \href {http://proceedings.mlr.press/v80/ren18a.html} {Learning to
  reweight examples for robust deep learning}.
\newblock In \emph{Proceedings of the 35th International Conference on Machine
  Learning, {ICML} 2018, Stockholmsm{\"{a}}ssan, Stockholm, Sweden, July 10-15,
  2018}, volume~80 of \emph{Proceedings of Machine Learning Research}, pages
  4331--4340. {PMLR}.

\bibitem[{Ren et~al.(2020)Ren, Li, Su, Kartchner, Mitchell, and
  Zhang}]{denoise_2020}
Wendi Ren, Yinghao Li, Hanting Su, David Kartchner, Cassie Mitchell, and Chao
  Zhang. 2020.
\newblock \href {https://doi.org/10.18653/v1/2020.findings-emnlp.334}
  {Denoising multi-source weak supervision for neural text classification}.
\newblock In \emph{Findings of the Association for Computational Linguistics:
  {EMNLP} 2020, Online Event, 16-20 November 2020}, volume {EMNLP} 2020 of
  \emph{Findings of {ACL}}, pages 3739--3754. Association for Computational
  Linguistics.

\bibitem[{Sang and De~Meulder(2003)}]{conll03_sang2003introduction}
Erik~F Sang and Fien De~Meulder. 2003.
\newblock Introduction to the conll-2003 shared task: Language-independent
  named entity recognition.
\newblock \emph{arXiv preprint cs/0306050}.

\bibitem[{Shu et~al.(2019)Shu, Xie, Yi, Zhao, Zhou, Xu, and
  Meng}]{meta_weight19}
Jun Shu, Qi~Xie, Lixuan Yi, Qian Zhao, Sanping Zhou, Zongben Xu, and Deyu Meng.
  2019.
\newblock \href
  {https://proceedings.neurips.cc/paper/2019/hash/e58cc5ca94270acaceed13bc82dfedf7-Abstract.html}
  {Meta-weight-net: Learning an explicit mapping for sample weighting}.
\newblock In \emph{Advances in Neural Information Processing Systems 32: Annual
  Conference on Neural Information Processing Systems 2019, NeurIPS 2019,
  December 8-14, 2019, Vancouver, BC, Canada}, pages 1917--1928.

\bibitem[{Van~der Maaten and Hinton(2008)}]{van2008visualizing_tsne}
Laurens Van~der Maaten and Geoffrey Hinton. 2008.
\newblock Visualizing data using t-sne.
\newblock \emph{Journal of machine learning research}, 9(11).

\bibitem[{Wang et~al.(2020)Wang, Mukherjee, Chu, Tu, Wu, Gao, and
  Awadallah}]{wang2020adaptive}
Yaqing Wang, Subhabrata Mukherjee, Haoda Chu, Yuancheng Tu, Ming Wu, Jing Gao,
  and Ahmed~Hassan Awadallah. 2020.
\newblock Adaptive self-training for few-shot neural sequence labeling.
\newblock \emph{arXiv preprint arXiv:2010.03680}.

\bibitem[{Xie et~al.(2020)Xie, Luong, Hovy, and Le}]{xieLHL20}
Qizhe Xie, Minh{-}Thang Luong, Eduard~H. Hovy, and Quoc~V. Le. 2020.
\newblock \href {https://doi.org/10.1109/CVPR42600.2020.01070} {Self-training
  with noisy student improves imagenet classification}.
\newblock In \emph{2020 {IEEE/CVF} Conference on Computer Vision and Pattern
  Recognition, {CVPR} 2020, Seattle, WA, USA, June 13-19, 2020}, pages
  10684--10695. Computer Vision Foundation / {IEEE}.

\bibitem[{Yarowsky(1995)}]{yarowsky95_selftrain}
David Yarowsky. 1995.
\newblock \href {https://doi.org/10.3115/981658.981684} {Unsupervised word
  sense disambiguation rivaling supervised methods}.
\newblock In \emph{33rd Annual Meeting of the Association for Computational
  Linguistics, 26-30 June 1995, MIT, Cambridge, Massachusetts, USA,
  Proceedings}, pages 189--196. Morgan Kaufmann Publishers / {ACL}.

\bibitem[{Yu et~al.(2021)Yu, Zuo, Jiang, Ren, Zhao, and Zhang}]{cosine2021}
Yue Yu, Simiao Zuo, Haoming Jiang, Wendi Ren, Tuo Zhao, and Chao Zhang. 2021.
\newblock \href {https://doi.org/10.18653/v1/2021.naacl-main.84} {Fine-tuning
  pre-trained language model with weak supervision: {A} contrastive-regularized
  self-training approach}.
\newblock In \emph{Proceedings of the 2021 Conference of the North American
  Chapter of the Association for Computational Linguistics: Human Language
  Technologies, {NAACL-HLT} 2021, Online, June 6-11, 2021}, pages 1063--1077.
  Association for Computational Linguistics.

\bibitem[{Zhang et~al.(2017)Zhang, Bengio, Hardt, Recht, and
  Vinyals}]{DBLP:conf/iclr/ZhangBHRV17}
Chiyuan Zhang, Samy Bengio, Moritz Hardt, Benjamin Recht, and Oriol Vinyals.
  2017.
\newblock \href {https://openreview.net/forum?id=Sy8gdB9xx} {Understanding deep
  learning requires rethinking generalization}.
\newblock In \emph{5th International Conference on Learning Representations,
  {ICLR} 2017, Toulon, France, April 24-26, 2017, Conference Track
  Proceedings}. OpenReview.net.

\bibitem[{Zhang et~al.(2021)Zhang, Yu, Li, Wang, Yang, Yang, and
  Ratner}]{wrench_benchmark_2021}
Jieyu Zhang, Yue Yu, Yinghao Li, Yujing Wang, Yaming Yang, Mao Yang, and
  Alexander Ratner. 2021.
\newblock \href {https://openreview.net/forum?id=Q9SKS5k8io} {{WRENCH}: A
  comprehensive benchmark for weak supervision}.
\newblock In \emph{Thirty-fifth Conference on Neural Information Processing
  Systems Datasets and Benchmarks Track}.

\bibitem[{Zhang et~al.(2015)Zhang, Zhao, and LeCun}]{agnews_zhang2015character}
Xiang Zhang, Junbo Zhao, and Yann LeCun. 2015.
\newblock Character-level convolutional networks for text classification.
\newblock \emph{Advances in neural information processing systems}, 28.

\bibitem[{Zhou et~al.(2022)Zhou, Xu, and McAuley}]{zhou2022meta}
Wangchunshu Zhou, Canwen Xu, and Julian McAuley. 2022.
\newblock Bert learns to teach: Knowledge distillation with meta learning.
\newblock In \emph{Proceedings of the 60th Annual Meeting of the Association
  for Computational Linguistics}, Online. Association for Computational
  Linguistics.

\bibitem[{Zhou et~al.(2020)Zhou, Lin, Lin, Wang, Du, Neves, and
  Ren}]{zhou2020_nero}
Wenxuan Zhou, Hongtao Lin, Bill~Yuchen Lin, Ziqi Wang, Junyi Du, Leonardo
  Neves, and Xiang Ren. 2020.
\newblock \href {https://doi.org/10.1145/3366423.3380282} {{NERO:} {A} neural
  rule grounding framework for label-efficient relation extraction}.
\newblock In \emph{{WWW} '20: The Web Conference 2020, Taipei, Taiwan, April
  20-24, 2020}, pages 2166--2176. {ACM} / {IW3C2}.

\bibitem[{Zhu et~al.(2022)Zhu, Hedderich, Zhai, Adelani, and
  Klakow}]{DBLP:conf/acl-insights/ZhuHZAK22}
Dawei Zhu, Michael~A. Hedderich, Fangzhou Zhai, David~Ifeoluwa Adelani, and
  Dietrich Klakow. 2022.
\newblock \href {https://doi.org/10.18653/v1/2022.insights-1.8} {Is {BERT}
  robust to label noise? {A} study on learning with noisy labels in text
  classification}.
\newblock In \emph{Proceedings of the Third Workshop on Insights from Negative
  Results in NLP, Insights@ACL 2022, Dublin, Ireland, May 26, 2022}, pages
  62--67. Association for Computational Linguistics.

\bibitem[{Zoph et~al.(2020)Zoph, Ghiasi, Lin, Cui, Liu, Cubuk, and
  Le}]{zophGLCLC020}
Barret Zoph, Golnaz Ghiasi, Tsung{-}Yi Lin, Yin Cui, Hanxiao Liu, Ekin~Dogus
  Cubuk, and Quoc Le. 2020.
\newblock \href
  {https://proceedings.neurips.cc/paper/2020/hash/27e9661e033a73a6ad8cefcde965c54d-Abstract.html}
  {Rethinking pre-training and self-training}.
\newblock In \emph{Advances in Neural Information Processing Systems 33: Annual
  Conference on Neural Information Processing Systems 2020, NeurIPS 2020,
  December 6-12, 2020, virtual}.

\end{thebibliography}
\bibliographystyle{acl_natbib}

\appendix

\clearpage
\section{Dataset Details}
\label{sec:appendix_dataset_details}
We experiment with eight NLP datasets, including six English datasets and two datasets in low-resource languages. All datasets come with their ground truth annotations and as well as the weak labels.

\subsection{Datasets Selection Criteria}
The WRENCH \cite{wrench_benchmark_2021} benchmark contains 23 NLP datasets. We choose representative datasets (like previous research in weak supervision) that \textbf{a)} overlap with previous works to enable direct comparisons. \textbf{b)} are diverse in terms of weak label quality, languages and tasks to approve the applicability of different baselines.

\subsection{English Datasets}

We experiment with four popular sequence classification datasets: AGNews, IMDB, Yelp and TREC.
\begin{enumerate}
  \item \textbf{AGNews} \cite{agnews_zhang2015character}: originates from AG, which is a large collection of news articles. The news are categorized in four classes: “World”, “Sports”, “Business” and “Sci/Tech”. 
  \item \textbf{IMDB} \cite{imdb_maas-etal-2011-learning}: consists of movie reviews with binary labels. It is a commonly used benchmark dataset for sentiment analysis.
  \item \textbf{Yelp} \cite{agnews_zhang2015character}: obtained from the Yelp Dataset Challenge in 2015. Similar to IMDB, it is a sentiment analysis dataset.
  \item \textbf{TREC} \cite{trec_li2002learning}: categorizes the questions in TREC-6 datasets into 6 categories: “Abbreviation”, “Entity”, “Description”, “Human”, “Location”, “Numeric-value”.
\end{enumerate}
and with the two sequence labeling datasets: CoNLL-03 and OntoNotes 5.0.
 \begin{enumerate}
   \item \textbf{CoNLL-03} \cite{conll03_sang2003introduction} NER dataset with four named-entity categories. 
  \item \textbf{OntoNotes 5.0} \cite{ontonotes_pradhan-etal-2013-towards}: NER dataset with 18 named-entity categories. 
 \end{enumerate}

All weak labels are obtained from the WRENCH benchmark\footnote{\url{https://github.com/JieyuZ2/wrench}} \cite{wrench_benchmark_2021}.

\subsection{Datasets in Low-Resource Languages}
Most datasets in the current WRENCH benchmarks are in English. Although weak supervision is desired in low-resource languages, it is understudied as finding annotators for them is more difficult. Hence, we include two low-resource languages, \yoruba and Hausa, to cover this scenario. Often, learning with weak labels in low-resource languages is more challenging. First, the training set is often much smaller than English datasets. For example, Hausa has only about 2k training samples while AGNews have 96k. Second, the weak labels in low-resource languages can have lower quality as experts for weak source development are harder to find. A set of simple rules is often used for labeling (which is the case in \yoruba and Hausa). Hence, weak supervision with low-resource languages is a combination of two challenges: training with \textit{small} datasets which have \textit{low-quality} labels.

\yoruba and Hausa are text classification datasets obtained from \cite{DBLP:conf/emnlp/HedderichAZAMK20}.\footnote{\url{https://github.com/uds-lsv/transfer-distant-transformer-african}}
\begin{enumerate}
    \item \textbf{\yoruba}: consists of news headlines from BBC Yoruba which are categorized in seven classes:  “Nigeria”, “Africa”, “World”, “Entertainment”, “Health”, “Sport”, “Politics”.
    \item \textbf{Hausa}: consists of news headlines from VOA Hausa which have the same seven classes as \yoruba. However, only five classes are considered in the text classification task. “Entertainment” and “Sport” have been removed due to the lack of samples of these classes.
\end{enumerate}

\citet{DBLP:conf/emnlp/HedderichAZAMK20} provided both the clean labels and weak labels on the two datasets. A gazetteer is created for each class for weak supervision. For example, a gazetteer containing names of agencies, organizations, states and cities in Nigeria is used to label the class “Nigeria”.

\begin{table*}[h]
\centering
\resizebox{\textwidth}{!}{
	\begin{tabular}{ccccccccc}
		\toprule \bf Dataset & \bf Task & \bf \# Class &\bf $|\mathcal{D}_w|$ &\bf $|\mathcal{D}_a|$ &\bf Coverage &\bf Conflict & \bf  $|\mathcal{D}_v|$ & \bf $|\mathcal{D}_t|$ \\ \midrule
        AGNews &Topic &4 &66,314 &96,000 &69.08\%  &14.17\% & 12,000 & 12,000  \\ 
        IMDB &Sentiment &2 &17,515 &20,000 &87.58\% & 11.96\% & 2,500 & 2,500 \\
        Yelp &Sentiment& 2 &25,165 &30,400 & 82.78\% & 18.29\% & 3,800 & 3,800  \\
        TREC & Question & 6 & 4,723 & 4,965 & 95.13\% & 22.76\% & 500 & 500   \\
        \yoruba &Topic &7 &1,340 &1,340 & 100.00\%  & 1.87\% & 189 & 379  \\
        Hausa &Topic &5 &2,045 &2,045 & 100.00\% & 1.90\% & 290 & 582  \\
        CoNLL03 & NER & 4 & 14,041 & 14,041 & 100.00\% & 4.05\% & 3,250 & 3,453     \\
        OntoNotes5.0 & NER & 18 &  115,812 &  115,812 & 100.00\% & 1.86\% & 5,000 &  22,897  \\
        \bottomrule
	\end{tabular}
}%
\caption{Dataset statistics. $|\mathcal{D}_w|$: number of training samples with weak labels. $|\mathcal{D}_a|$: total number of training samples (weakly labeled + unlabeled). Coverage: fraction of samples that are weakly labeled, i.e., $\frac{|\mathcal{D}_w|}{|\mathcal{D}_a|}$. Conflict: samples that are labeled by at least two weak sources with contradicted weak labels. $|\mathcal{D}_v|$: number of validation samples. $|\mathcal{D}_t|$: number of test samples.}
\label{tab:dataset_appendix}

\end{table*}

\begin{table*}[h!]
    \centering
    \begin{tabular}{lc}
        \toprule 
        \textbf{Hyperparameter} & \textbf{Search Range} \\\midrule  
         Teacher Learning Rate & 3e-6, 5e-6, 2e-5, 3e-5\\
         Teacher Warm-Up Steps & 500, 100, 2000, 3000\\
         Confidence Filter Threshold  &  0.4, 0.5, 0.6, 0.7, 0.8, 0.95\\
         \bottomrule 
    \end{tabular}
    \caption{Hyperparameter search.}
    \label{tab:hyperparameters_search}
\end{table*}

\subsection{More Dataset Statistics}
We provide more details on the datasets we used in our experiments in Table \ref{tab:dataset_appendix}. In general, not all data can be covered by weak sources. No weak source is triggered for some training samples and they remain unlabeled. The coverage of the datasets ranges from 69.08\% to 100\%. Note that for NER tasks, the coverage is always 100\% since if no weak source is triggered for a token, we assign label ``O'' (i.e., non-entity) to it. On the other hand, some samples can be covered by two or more weak sources with contradicted weak labels. In this case, we have a conflict. The conflict ratio ranges from 1.86\% to 22.76\% in the datasets we tested.

\begin{table*}[h!]
\centering
\resizebox{2\columnwidth}{!}{
\begin{tabular}{lcccccccc}
\toprule  
& \textbf{AGNews}  & \textbf{IMDB}   & \textbf{Yelp} & \textbf{TREC} & \textbf{\yoruba}  & \textbf{Hausa} &  \textbf{CoNLL-03} &  \textbf{OntoNotes 5.0}  \\\midrule 

BERT Backbone & RoBERTa & RoBERTa  & RoBERTa &  RoBERTa & mBERT  & mBERT & RoBERTa & RoBERTa\\
Batch Size & 32 & 16  & 16 & 32 & 32 & 32 & 32 & 32 \\
Maximum Sequence Length & 128 & 256 & 256 & 64 & 64 & 128 & 64 & 64\\
Student Learning Rate & 2e-5 & 2e-5 & 2e-5 & 2e-5 & 2e-5 & 2e-5 & 2e-5 & 2e-5\\
Teacher Learning Rate & 2e-5 & 2e-5 & 2e-5 & 2e-5 & 5e-6 & 2e-5 & 2e-5 & 2e-5\\
Teacher Warm-Up Steps & 500 & 500  & 3000 & 500 & 1000 & 3000 & 2000 & 2000 \\
Confidence Filter Threshold & 0.7 & 0.7 & 0.5 & 0.5 & 0.7 & 0.4 & 0.8 & 0.5\\
\bottomrule 
\end{tabular}}
\caption{Selected hyperparameters. mBERT: multilingual BERT.}
\label{tab:best-hyperparameters}
\end{table*}

\section{Implementation Details}
\label{sec:appendix_implementation_details}

\paragraph{Models.} All baselines in our paper, except the majority vote and the Snokerl model \cite{snorkel_2017} which work with label space only, use the official RoBERTa model\footnote{\url{https://huggingface.co/roberta-base}} \cite{liu19roberta} from Huggingface as the classification backbone for all English datasets, and the multilingual BERT\footnote{\url{https://huggingface.co/bert-base-multilingual-cased}} for datasets in African languages. We use the base version of the two models which contain roughly 120M and 110M parameters, respectively.

\begin{table}[h!]
\centering
\resizebox{0.80\columnwidth}{!}{
	\begin{tabular}{ccc}
		\toprule \bf Dataset & \bf Test & \bf Validation  \\ \midrule
        AGNews & 89.92 & 89.90 \\ %
        IMDB & 89.16 & 89.21  \\ %
        Yelp & 95.00 & 94.79  \\ %
        TREC & 94.80 & 94.42  \\ %
        \yoruba & 72.56 & 75.13  \\ %
        Hausa & 59.11 & 62.34  \\ %
        CoNLL-03 & 88.41 & 87.86  \\ %
        OntoNotes & 74.59 & 75.20  \\ %
        \bottomrule
	\end{tabular}
}%
\caption{The average test and validation accuracy/F1 score (in \%) of MSR over five trials.}
\label{tab:validation_performance_appendix}
\end{table}

\begin{table*}[h!]
\centering
\resizebox{2\columnwidth}{!}{
\begin{tabular}{lcccccccc}
\toprule  
\textbf{MSR Configuration}     &  \begin{tabular}[c]{@{}c@{}} \textbf{AGNews}\\ \textbf{(Acc)} \end{tabular} &  \begin{tabular}[c]{@{}c@{}} \textbf{IMDB}\\ \textbf{(Acc)} \end{tabular} &  \begin{tabular}[c]{@{}c@{}} \textbf{Yelp}\\ \textbf{(Acc)} \end{tabular} &  \begin{tabular}[c]{@{}c@{}} \textbf{TREC}\\ \textbf{(Acc)} \end{tabular} &  \begin{tabular}[c]{@{}c@{}} \textbf{\yoruba}\\ \textbf{(Acc)} \end{tabular} &  \begin{tabular}[c]{@{}c@{}} \textbf{Hausa}\\ \textbf{(Acc)} \end{tabular} &  \begin{tabular}[c]{@{}c@{}} \textbf{CoNLL-03}\\ \textbf{(F1)} \end{tabular} &  \begin{tabular}[c]{@{}c@{}} \textbf{OntoNotes}\\ \textbf{(F1)} \end{tabular}   \\\midrule   
Student & \textbf{89.92} & \textbf{89.16} & \textbf{95.00} & \textbf{94.80}  & \textbf{72.56} & 59.11 & \textbf{88.41} & \textbf{74.59}    \\
Teacher & 89.02 & 88.08 & 94.37 & 93.80  & 68.87 & \textbf{60.14} & 87.30 & 73.22 \\
w/o Teacher Scheduler & 89.68 & 87.68 & 93.78 &	93.60 & 70.71 & 55.32 & 87.82 & 72.48 \\
w/o Confidence Filtering & 89.87 &	89.04 & 94.76 & 93.60 & 71.50 & 55.15  & 88.07 & 74.11 \\
w/o Both & 89.55 &	87.68 & 93.33 & 93.40 & 70.50 & 55.32 & 87.82 & 72.08 \\
\bottomrule 
\end{tabular}}
\caption{Ablation studies. The numbers represent the test accuracy and F1 Score.}
\label{tab:ablation-results-detailed}
\end{table*}

\begin{table*}[h!]
\centering\scriptsize
\begin{tabular}{@{}lcccccccc@{}}
\toprule
 & AGNews & IMDB & Yelp & TREC & \yoruba & Hausa & CoNLL-03 & OntoNotes 5.0 \\ \midrule
Running time (hours) & 2.5 & 1.6 & 0.5 & 1.2 & 0.5 & 0.7 & 1.1 & 3.0 \\ \bottomrule
\end{tabular}
\caption{Average runtime (in hours) for training a MSR model. One single Nvidia Tesla V100 GPU is used in each experiment to accelerate the computation.}
\label{tab:average_runtime}
\end{table*}

\paragraph{Fine-Tuning on Classification Task.} We fine-tune all layers using AdamW \cite{LoshchilovH19_adamW} as the optimizer. For sequence classification tasks, we pass the final layer of the [CLS] token representation ($\mathbb{R}^{768}$) to a feed-forward layer for prediction. For sequence labeling tasks, the final layers of all tokens ($\mathbb{R}^{768 \times L}$, where $L$ is the sentence length) are passed to a shared feed-forward layer to predict the class of each token in the sentence. We report the score where the model performs the best on the validation set during training.

\paragraph{Hyper-Parameters of MSR.}
We apply grid search on the warm-up steps for the teacher and the confidence threshold for the student network. Table \ref{tab:hyperparameters_search} shows our hyperparameter search configuration. We choose the final configurations of the hyperparameters according to the model's performance on the validation set. Table \ref{tab:best-hyperparameters} shows the best configurations of parameters we used to produce the results in Table \ref{tab:main_result}.

\paragraph{Evaluation Metrics.}
For model evaluation, we report accuracy for sequence classification tasks and F1 Score for sequence labeling tasks. In our implementation, we call the function \verb|classification_report()| from the scikit-learn library\footnote{\url{https://scikit-learn.org/stable/modules/generated/sklearn.metrics.classification_report.html}} to compute the accuracy, and use the \verb|Seqeval| class from Huggingface\footnote{\url{https://github.com/huggingface/datasets/blob/master/metrics/seqeval/seqeval.py}} to compute the F1 Score.

\section{Validation Performance}
The average test performance of MSR is reported in Table \ref{tab:main_result}. We further report the corresponding validation performance in Table \ref{tab:validation_performance_appendix}.

\section{Ablation Studies}
\label{sec:ablation_studies_detailed}
We report the detailed ablation results for each dataset in Table \ref{tab:ablation-results-detailed}.

\section{Hardware and Average Runtime.}
\label{appendix:averge_runtime_msr} 
We use Nvidia Tesla V100 to accelerate training. The average runtime for each method and dataset is summarized in Table \ref{tab:average_runtime}.

\end{document}